\documentclass{article}

\usepackage{PRIMEarxiv}

\usepackage[utf8]{inputenc} % allow utf-8 input
\usepackage[T1]{fontenc}    % use 8-bit T1 fonts
\usepackage{hyperref}       % hyperlinks
\usepackage{url}            % simple URL typesetting
\usepackage{booktabs}       % professional-quality tables
\usepackage{amsfonts}       % blackboard math symbols
\usepackage{nicefrac}       % compact symbols for 1/2, etc.
\usepackage{microtype}      % microtypography
\usepackage{lipsum}
\usepackage{fancyhdr}       % header
\usepackage{graphicx}       % graphics
\graphicspath{{media/}}     % organize your images and other figures under media/ folder

\usepackage{amsmath}
\usepackage{makecell}
\newtheorem{definition}{Definition}

\usepackage{caption}
\usepackage{subcaption}
\usepackage{multirow}
\usepackage{graphicx}
\usepackage{tikz}
\usepackage{natbib}

\title{Crossing Boundaries: Leveraging Semantic Divergences to Explore Cultural Novelty in Cooking Recipes
}

\author{
  Florian Carichon\thanks{This is the version of the paper accepted for publication at the ACM Conference on Fairness, Accountability, and Transparency (FAccT 2025).} \\
  MILA, McGill University \\
  Montréal, Canada\\
  \texttt{florian.carichon@mila.quebec} \\
   \And
  Romain Rampa \\
  École de technologie supérieure (ÉTS) \\
  Montréal, Canada\\
  \texttt{romain.rampa@etsmtl.ca} \\
  \And 
  Golnoosh Farnadi \\
  MILA, McGill University \\
  Montréal, Canada\\
  \texttt{farnadig@mila.quebec}
}

\begin{document}
\maketitle

\begin{abstract}
Novelty modeling and detection is a core topic in Natural Language Processing (NLP), central to numerous tasks such as recommender systems and automatic summarization. It involves identifying pieces of text that deviate in some way from previously known information. However, novelty is also a crucial determinant of the unique perception of relevance and quality of an experience, as it rests upon each individual's understanding of the world. Social factors, particularly cultural background, profoundly influence perceptions of novelty and innovation. Cultural novelty arises from differences in salience and novelty as shaped by the distance between distinct communities. While cultural diversity has garnered increasing attention in artificial intelligence (AI), the lack of robust metrics for quantifying cultural novelty hinders a deeper understanding of these divergences. This gap limits quantifying and understanding cultural differences within computational frameworks. To address this, we propose an interdisciplinary framework that integrates knowledge from sociology and management. Central to our approach is GlobalFusion, a novel dataset comprising 500 dishes and approximately 100,000 cooking recipes capturing cultural adaptation from over 150 countries. By introducing a set of Jensen-Shannon Divergence metrics for novelty, we leverage this dataset to analyze textual divergences when recipes from one community are modified by another with a different cultural background. The results reveal significant correlations between our cultural novelty metrics and established cultural measures based on linguistic, religious, and geographical distances. Our findings highlight the potential of our framework to advance the understanding and measurement of cultural diversity in AI.
\end{abstract}

\section{Introduction}
\label{sec:intro}

The study of novelty modeling and measurement has long been a pivotal topic in the NLP community, which focuses on identifying relevant information that is unknown relative to a predefined state of knowledge \citep{priarone2024unsupervised}. Novelty detection is fundamental for various applications, including automatic summarization, search engines, and recommender systems \citep{mohseni2022framework}. While numerous studies have explored modeling approaches \citep{mohseni2022framework} and algorithmic techniques \citep{pimentel2014review}, relatively few have examined how novelty detection relates to information needs, which depend on the definition of a user’s pre-existing knowledge \citep{lavrenko2008generative}. Novelty itself is inherently multifaceted, interpreted differently by individuals, groups, societies, or historical contexts \citep{dean2006identifying}. Factors such as familiarity, diversity, complexity \citep{chersoni2021not}, and symmetric or asymmetric perceptions \citep{tsai2010evaluation} further shape its interpretation.

A user group's perception of the quality of information and search results is deeply influenced by \emph{culture} \citep{komlodi2004identifying}, a shared system of meanings that defines values, norms, and beliefs \citep{paletz2008implicit}. Consequently, what is meaningful and useful to one culture may be irrelevant or incomprehensible to another \citep{kim2013information}. The concept of \emph{cultural novelty}, rooted in sociology and management, captures the subjective perception of familiarity or difference within a cultural context \citep{wilczewski2023cultural}. Such differences become particularly evident in culturally anchored products like food \citep{guerrero2009consumer}, where relevance, innovation, and salience vary with cultural origin \citep{goh2009culture} and interactions across cultural distances \citep{paletz2008implicit}.

In the context of NLP and AI, the emergence of generative large language models (LLMs) has spurred research into cultural diversity and awareness, primarily through external tasks involving prompting \citep{bhatt2024extrinsic,li2024culturellm}. Among such approaches, researchers try to measure LLMs personality traits, values and cultural perspective through self-assessment tests and controllable prompts \citep{kovač2023largelanguagemodelssuperpositions}. Recent work using the WVS Inglehart–Welzel cultural map \citep{WVSCultMap} proposed interpretable cultural distance metrics to evaluate LLM outputs \citep{tao2024cultural}. While this approach highlights cultural representation differences, its reliance on specific questionnaires limits generalizability across tasks and datasets. In this paper, we propose five new \emph{cultural novelty} detection metrics based on probability distribution divergences, an area of interest in out-of-distribution detection \citep{yang2024generalized}, which exhibits promising properties, such as entropic uncertainty, when applied to LLMs \citep{ahmadian2024unsupervised}. In sociology and management, divergence measures have been widely used to analyze cultural evolution, legal systems, societal norms \citep{klingenstein2014civilizing}, and community values \citep{gallagher2018divergent}. This work bridges the gap between cultural novelty and distance in sociology and the challenges of cultural representation and diversity in AI. Leveraging divergence-based novelty detection, we propose new metrics to measure cultural differences through text analysis. By adapting these well-established textual novelty measures to the cultural domain, we aim to demonstrate their potential to meaningfully capture and generalize cultural differences across contexts, focusing on cultural proxies like food. Our contributions are as follows:

\begin{itemize}
\item Proposing five information theoretic \emph{cultural novelty} metrics which enable the measurement of various divergence phenomena in textual data (Section~\ref{sec:divergence})
\item Introducing \textit{GlobalFusion}, a new dataset that facilitates the comparison of text descriptions of the same cultural products across different cultures. The dataset studies 500 different dishes for a total of 135 thousand recipes spanning 173 different countries.
\item Evaluating our proposed metrics with four cultural distances: Inglehart–Welzel cultural distance, geographical distance, linguistic distance, and religious distance, derived from social studies. We examine how our proposed cultural novelty dimensions relate to cultural differences in the social science literature to bridge the gap between the two fields of AI and Sociology.
\end{itemize}

Our study proposes a new framework to evaluate cultural distance in textual productions, focusing on how humans adapt cultural products to their own cultures. Our results show that, in this framework, novelty divergence metrics correlate significantly to cultural distances and that the strength of correlation depends on the studied cultural aspect. The benefit of this framework is that it is easily adaptable to different tasks and cultural proxies, enabling extensive evaluation of textual productions that will enable a potential future use for understanding and enhancing of LLMs' cultural representative capacity.

\section{Related work}

This work builds on foundational sociological studies of cultural perception, distance, and novelty, as well as advancements in NLP and AI novelty detection, which we outline in the remainder of this section.

\subsection{Sociological Perspectives on Cultural Novelty}
\label{sec:related_cult}

Culture defines common ground knowledge, aboutness, and the values, norms, and beliefs of a group of people \citep{paletz2008implicit, hershcovich2022challenges}. Therefore, communities that do not share all these fundamental principles perceive information differently and what features are meaningful for their comprehension of the world \citep{kim2013information}. For example, sociological studies have highlighted that the relevance of new information, salient features, and how innovation is valued heavily depend on people's culture of origin \citep{goh2009culture} and the distance to the culture of others and their interactions \citep{paletz2008implicit}. Finally, researchers have highlighted that these differences become even more evident when culturally anchored products such as food are evaluated \citep{guerrero2009consumer}. These differences represent a continuum that presumably measures the extent to which different cultures are similar or different, known as the concept of cultural distance \citep{shenkar2001cultural}. Since then, multiple distances have been proposed to measure national cultural distances, based on geographical distances to explain issues in corporation acquisition or student experience \citep{shenkar2001cultural, wilczewski2023cultural}; based on traditional and survival values known as the Inglehart–Welzel Cultural Map \citep{WVSCultMap}; and even based the differences between the major languages\footnote{\url{http://dow.net.au/?page_id=32}} or religion\footnote{\url{http://dow.net.au/?page_id=35}} practiced in countries \citep{gordon2005ethnologue}. While these bear an objective and physical concept, cultural distance can be intimately subjective when it represents an individual's experience that needs to adapt to a different culture \citep{liang2023emerging}. This new notion, termed cultural novelty, relates to the perceived divergence between one culture and another \citep{wilczewski2023cultural}. In social science, several studies have proposed relying on communication styles, linguistic norms, and style to measure cultural differences between groups. \cite{pechenick2015characterizing} used relative entropy to measure socio-cultural evolution in book corpora. \cite{klingenstein2014civilizing} studied how language diverges and evolves in time with the creation of new social laws and norms. Other authors again used relative entropy to measure the divergence of values in two different social groups \citep{gallagher2018divergent}. In the artificial intelligence community, there is a need to come up with tools that will allow understanding of the shared knowledge and assumptions among people, known as common ground knowledge and the distinct way of "aboutness" or prioritizing different topics as relevant in various communities \citep{hershcovich2022challenges}. An interesting recent work proposed by \cite{tao2024cultural} proposed to evaluate LLMs output based on the WWS's Inglehart–Welzel Cultural Map. Our study introduces a novel evaluation framework for the artificial intelligence community, featuring a new dataset and accompanying metrics derived from natural language processing and distribution divergence, grounded in theories from sociology and management.

\subsection{Computational Linguistic Methods for Novelty Detection}
\label{sec:related_ood}

Novelty detection is finding new or unknown information that diverges significantly from expected or established patterns and identifying items that differ from all other items or that would be unknown or unexpected to a user \citep{priarone2024unsupervised}. More specifically, novelty can be viewed from different angles: a novel item is an item that is not similar to a prototypical view of its fields, not similar to every other item observed, that brings new features or combinations of features never observed before, or finally as data that contain more information for the observer \citep{mohseni2022framework}. Multiple approaches exist relying on analyzing fundamental novelty properties such as frequency of events, Bayesian surprise, information theory, extreme value statistics, or kernel approaches \citep{pimentel2014review}. Recently, novelty detection has been grouped with anomaly detection, open set recognition, and in a general framework under the topic of Out-of-Distribution Detection (OOD) \citep{yang2024generalized}. Indeed, for all these tasks, the central point is to define the normal distribution, generally drawn from the data available during training and to detect points that diverge from that normality. Due to its use in social studies, we take an interest in information theory-based approaches that aim to understand semantic novelty between texts. As we said, novelty refers to a multiplicity of phenomena that can be connected to different aspects of distribution divergence \citep{boult2020unifying}. First, approaches rely on non-symmetrical definition of novelty equivalent to frequency \citep{tsai2010evaluation}. \cite{he2005optimization} considered novel data points with the highest entropy. This idea was also used to detect which features contribute the most to the said novelty \citep{gamon2006graph}. More recently, some authors have define the ideal highest point of entropy to be considered as a valuable and acceptable novelty \citep{miyamoto2020modeling}. Other authors instead used a symmetric approach based on distances. \cite{ando2007clustering} employed an equivalent clustering approach to compare the distribution of new points with centroids distribution. The authors in \citep{gabrilovich2004newsjunkie} followed a more k-NN-oriented approach by comparing the distribution of new information with the distribution of all observed news seen before. Finally, the authors used an atypical combination to define prior distribution in the Bayesian surprise \citep{baldi2010bits,itti2005bayesian} framework to detect novelty and surprise \citep{grace2017encouraging,brown2015computational,grace2015data}. Our study proposes to define the normal distribution with a cultural aspect and to use information-theory-based approaches to detect how new points coming from different culture diverge from our defined normality.

\section{Methodology}
\label{sec:metho}

We aim to propose efficient methods for measuring cultural differences by analyzing language novelty and divergences in depicting cultural proxies. The remainder of the section introduces the necessary conditions for the application of these divergence methods to evaluate the notion of cultural novelty. 

%The notion of novelty refers to a multiplicity of phenomena connected to different information-theoretical definitions that we regroup under four distinct concepts that we name: (1) Newness, (2) Uniqueness, (3) Difference, (4) Surprise. The remainder of the section introduces the necessary conditions for the application of these divergence methods to evaluate the notion of cultural novelty. 

\subsection{Divergence Metrics}
\label{sec:divergence}

To estimate the difference in information content between our two distributions, we rely on the Jensen-Shannon divergence (JSD). JSD is a reformulation of the well-known Kullback-Liebler (KL) divergence to make it a symmetric distance bounded between 0 and 1 \citep{lin1991divergence} and is expressed as:

\begin{equation}
    D_{JS}(P||Q) = \pi_1 D_{KL}(P||M) + \pi_2 D_{KL}(Q||M)
\end{equation}

where $D_{KL}$represents the KL divergence, $P$ is the training distribution, $Q$ is the test distribution and $M$ is the mixed distribution $M = \pi_1P + \pi_2Q$ and $\pi_1$, $\pi_2$ are weights proportional to the sizes of $P$ and $Q$ such that $\pi_1 + \pi_2 = 1$.

\subsection{Culturally oriented knowledge space}
\label{sec:KBEB}

\begin{figure}
    \centering 
    \includegraphics[width=0.8\textwidth , trim={10mm 55mm 30mm 5mm}]{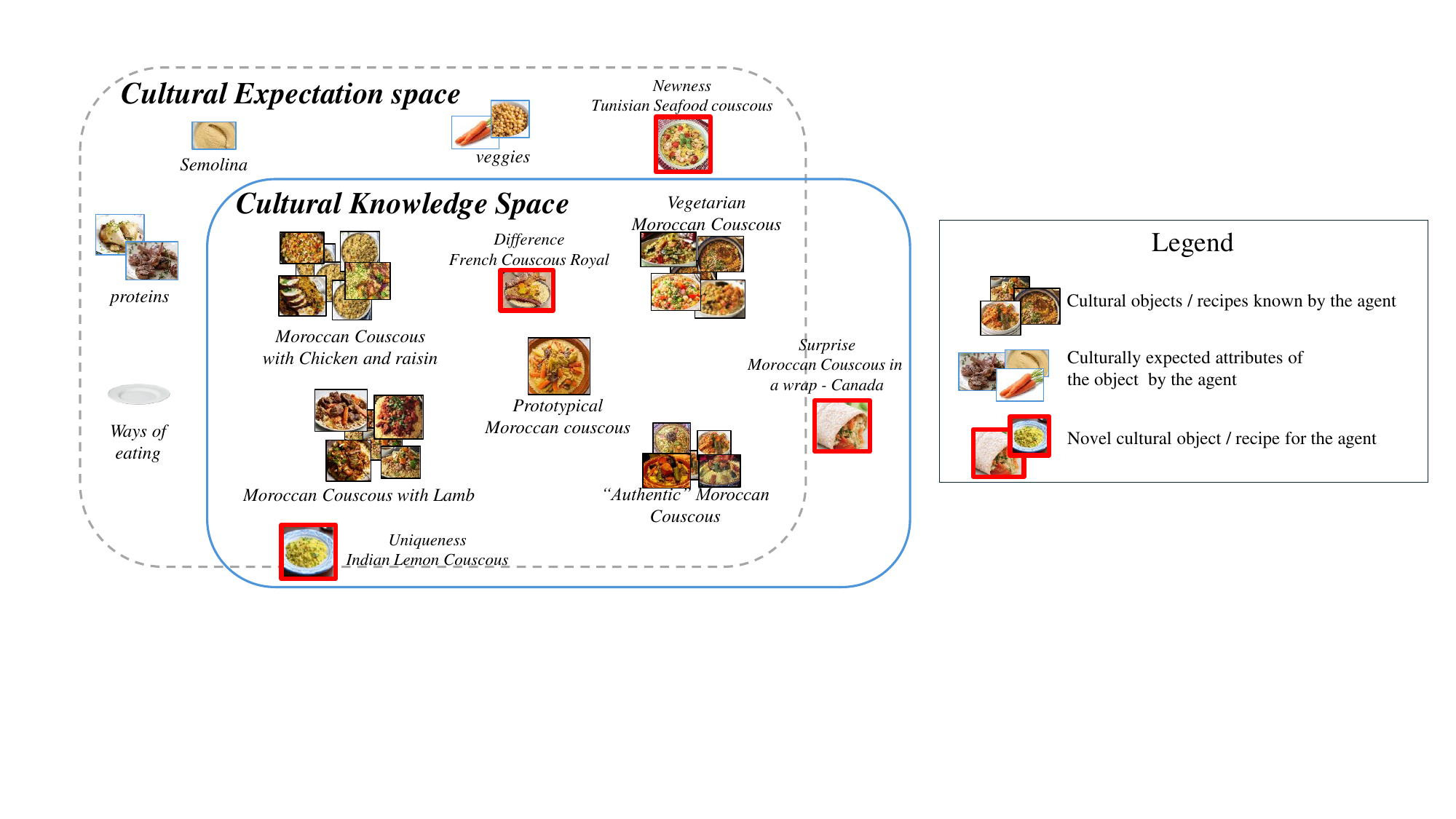}
    \caption{Example of a knowledge and expectation of an agent representation of a Moroccan Couscous. Different types of novel observation can be observed and relate differently to the agent space.}
    %\Description{Hand made graphic representing knowledge and expectation spaces.}
    \label{fig:KBEB}
\end{figure}

The central point with novelty detection is the definition of the knowledge state of the agent that will observe the novel event \citep{yang2024generalized, boult2020unifying}, that will define $P$. In novelty detection, this knowledge refers to the observed distribution extracted from the known corpus of documents in our database \citep{pimentel2014review}. This knowledge space will also establish an expectation space related to this corpus, which will relate to its possible evolutions and not only to its attributes but to its likely combinations and recombination \citep{grace2017encouraging}. Figure \ref{fig:KBEB} illustrates these knowledge and expectation spaces and how novel observations are positioned with respect to them. 

To include a cultural dimension, we need to define the knowledge space and the documents in such a way that they represent cultural aspects. Since our metrics focus on measuring textual divergence, the inclusion of this cultural dimension is mandatory for defining cultural novelty.  The corpus must respect two basic conditions: (1) the text must depict products that embedded cultural representations, and (2) it possesses an explicit variable linking the product to a cultural community. Therefore, we must have a set of products $Prod$ acting as artifacts of a country's culture (food, painting, movies, literature ). We also have to define a set of variables $C$ that represent a set of cultural attributes related to the product (country of origin, religion, author's genre). Finally, in our context, we need a set of texts $T$ that can be considered a relevant depiction of these cultural products. Once these conditions are fulfilled, we can define our cultural knowledge space as the set of all texts $T_{prod_k;c_j}$ associated with the cultural product $prod_k$ and the cultural variable $c_j$. We define $P$ as the normal distribution, learned from our knowledge space, and the new distribution $Q$, associated with our new product. We can then express $P$ and $Q$ as : 
\begin{itemize}
    \item $P_{prod_k;c_j} = P(w|T_{prod_k;c_j})$ is the probability distribution of the set of documents of a given cultural product $prod_k$ and a given cultural variable $c_j$.
    \item $Q_{prod_k} = P(w|NT_{prod_k;c_i})$ is the probability distribution of a new document $NT$ of the same cultural product $prod_k$ and a given cultural variable $c_i$.
\end{itemize}
 
We can now introduce different novelty methods, which we name (1) newness, (2) uniqueness, (3) difference, and (4) surprise, to measure divergence to the cultural knowledge base.

\subsubsection{Newness}
\label{sec:new}

One of the elements most frequently associated with novelty is that of newness, to describe the appearance or disappearance of properties, terms or knowledge that had never been observed in the initial knowledge base \citep{grace2015data, miyamoto2020modeling}. We consider large or small relative entropy values in the new distribution to detect the most relevant features for novelty \citep{gamon2006graph}. Once a detection threshold is determined, any individual term that exceeds the threshold is considered a significant deviation \citep{pimentel2014review}. More specifically, we use the linear properties of the JSD to extract the contribution of a word to the JSD: 

\begin{equation}
     D_{JS,i}(P{prod_k;c_j}||Q_{prod_k;c_i}) = -m_i \log_2(m_i) + \pi_1 p_i \log_2(p_i) + \pi_2 q_i \log_2(q_i)
\end{equation}

where $m_i$ is the probability of seeing word i in $M$. The contribution from word $i$ is 0 if and only if $p_i = q_i$. Therefore, we can label the contribution to the divergence from the word $i$ as coming from $Q_{prod_k;c_i}$ by determining if $q_i$ is greater than $p_i$, beyond a certain threshold \citep{gallagher2018divergent}. 

\begin{definition}
\textit{Cultural Newness}: The proportion of words significantly appearing or disappearing in the distribution of the text coming from a cultural variation compared to the distribution of our cultural knowledge base.
\end{definition}

\begin{equation}
    N_{x} = \frac{1}{N_{Q_{prod_k;c_i}}}\sum_{i=1}^{N_{Q_{prod_k;c_i}}} \delta_i \quad \text{with} \quad
    \delta_i = 
    \begin{cases}
       1 & \text{if } x_i \geq \epsilon, \\
       0 & \text{if } x_i \leq \epsilon,
    \end{cases}
\end{equation}

Where \(N_x\) represents either \textit{Appearance} or \textit{Disappearance} depending on the choice of \( x \). If \( x = q \), the equation calculates \textit{Appearance}. If \( x = p \), the equation calculates \textit{Disappearance}. $N_{Q^{ij}_{p_k}}$ is the number of words in the new text $NT_{prod_k}$, or the number of quantitative variables in the distribution $Q_{prod_k;c_i}$, and $\epsilon$ is the newness threshold. We employ a leave-one-out strategy in our cultural knowledge base to set our newness threshold. For each text $t$ in the set $T_{prod_k;c_j}$ representing the cultural product $Prod_k$, we measure its divergence to the set without that document. This strategy allows us to represent the average term appearance inside the same community, meaning that to be relevant, a new term should contribute to the divergence more than it is observed for a cultural product associated with a specific community. Finally, we can estimate the \textit{cultural newness} score as:

\begin{equation}
    Newness = \lambda_1 Appearance + \lambda_2 Disappearance
\end{equation}

Where $\lambda_1$ and $\lambda_2$ are weighting factors to account more for the appearance of new terms or the disappearance of old terms. We have set to $\lambda_1 = 0.8$ and $\lambda_2 = 0.2$ in our study to favor the appearance of the new term.

\subsubsection{Uniqueness}
\label{sec:uniq}

The concept of uniqueness refers to the uncommon aspect of an object, given that its characteristics are somewhat dissimilar from an agent's view of the knowledge space  \citep{greul2023does, boult2020unifying}. There is thus the idea of a certain distance from the identity of an object or a prototypical vision of it \citep{rosch1975cognitive}. In the novelty detection framework, this concept strongly connects to clustering-based approaches, where a small number of prototype points in the data space characterizes the "normal" class \citep{mohseni2022framework, pimentel2014review}. The minimum distance from a test point to the nearest prototype is often used to quantify its novelty and divergence \citep{ando2007clustering}.

\begin{definition}
\textit{Cultural Uniqueness}: The divergence measure of the distribution of a new text associated with a cultural variation compared to the prototypical view of the cultural product within a community. 
\end{definition}
In our case, $P_{prod_k;c_j}$ directly represents this average representation of our cultural product. The cultural uniqueness score is thus simply defined as:

\begin{equation}
    Uniqueness = D_{JS}(P_{prod_k;c_j}||Q_{prod_k;c_i})
\end{equation}

Therefore, the greater the divergence of a cultural product from this average representation, the greater the cultural uniqueness of the product.

\subsubsection{Difference}
\label{sec:diff}

Similarity-based approaches also involve applying a distance-based metric to find how similar a text document is with every other document constituting the knowledge space. Contrary to the concept of uniqueness, the concept of difference is measured by the degree to which a design differs from those that have come before it \citep{grace2015data}. Evaluation is done by comparison with descriptions of past or existing designs or objects \citep{gabrilovich2004newsjunkie}. This formalization links the notion of novelty to the hypothesis that people categorize items by comparing them to all examples stored in their memory \citep{nosofsky2011generalized}. This concept is concretely applied through nearest neighbor-based approaches, which suppose that normal data points have close neighbors in the "normal" training set while novel points are far from those points \citep{pimentel2014review, gabrilovich2004newsjunkie}.

\begin{definition}
\textit{Cultural Difference}: The distribution of a new text describing a cultural variation is distant from all the observed points in the knowledge base associated with that cultural product. 
\end{definition}

More specifically, in the neighborhood of the new point, we count the proportion of points that exceed a certain threshold distance. The more points there are beyond this single point, the more the point can be considered different from its neighborhood and, therefore, new. The formulation of the cultural difference score is as follows: 

\begin{equation}
    Difference = \frac{1}{|T_{prod_k;c_j}|} \sum_{m=1}^{|T_{prod_k;c_j}|} \delta_i with  
    \begin{cases}
       \delta_i =  1 & \text{if  }  D_{JS}(P_{t_m; p_k;c_j}||Q_{p_k;c_i}) \geq \epsilon, \\
       \delta_i =  0 & \text{if  }  D_{JS}(P_{t_m; p_k;c_j}||Q^{ij}_{p_k} \leq \epsilon, 
    \end{cases}
\end{equation}

Where $\epsilon$ is the threshold for difference and $P_{t_m; p_k;c_j}$ is the probability distribution of a specific text in the set $T_{prod_k;c_j}$ associated with a cultural product from a community $j$. As for newness, we define the threshold for difference through a leave-one-out strategy in our knowledge base. For each text $t_m$ in $T_{prod_k;c_j}$, we measure the distance to all the other documents present in $T_{prod_k;c_j}$ and average it for the whole community $j$. This strategy allows us to consider that a new document will be different if it is farther away than its neighbors than the community's average distance between points.

\subsubsection{Surprise}
\label{sec:surp}

The last concept of surprise stems from a violation of expectations in a space of projected  designs rather than in a space of existing designs \citep{maher2013computational}. The formalization of this concept is transcribed by Bayesian surprise \citep{baldi2010bits, itti2005bayesian} as a divergence between two distributions that relate to its likely combinations and recombination \citep{grace2017encouraging, grace2015data}. In language, this often corresponds their Pointwise Mutual Information (PMI) \citep{grace2017encouraging}. However, multiple concurrent approach relying either on frequency or divergence approaches co-exists in the literature to define surprise \citep{mohseni2022framework}.

\begin{definition}
\textit{Cultural Surprise}: The distribution of expected designs or combinations of attributes of a new cultural variation violates the projection of the cultural knowledge space distribution into a cultural expectation base. This violation can either be transcribed into \textit{New Surprise} as the appearance of new cultural combinations that did not exist in the cultural expectation space or into \textit{Divergent Surprise} as the absolute divergence of the distribution of combination of the cultural variation.
\end{definition}

The first metric \textit{Cultural New Surprise} relates to the shear appearance of new combinations that did not exist in the expectation space \citep{gamon2006graph, carayol2019right}. More formally, let us define the PMI distribution of each term, representing our prior distribution in the cultural expectation space, and the observed PMI distribution of the terms of a new text. Given $W_T$, the set of unique $M$ words present in $T_{prod_k;c_j}$, associated with a product $k$ and a community $j$; and $W_{NT}$, the set of $N$ words in the new text $NT_{prod_k;c_i}$ associated with the same cultural product but not necessarily the same community. Knowing that the MI between two words is defined as $MI(w_i,w_j) = \log \frac{p(w_i,w_j)}{p(w_i)p(w_j)}$, we can  create the representations of the PMI matrix $PPMI \in \mathbb{R}^{M*M}$ estimated by the co-occurrence of all terms in $W_T$ and $QPMI \in \mathbb{R}^{N*N}$ estimated from the co-occurrence of all terms in $W_{NT}$. 
\begin{equation}
    New Surprise = \frac{1}{N} \sum_{i,j}^{N}
    \begin{cases}
    1 & \text{if  } QPMI(w_i,w_j) > 0 \land PPMI(w_i,w_j) =0\\
    1 & \text{if  } {w_i,w_j} \notin PPMI \\
    0 & else
    \end{cases}
\end{equation}

The second notion \textit{Cultural Divergent Surprise} relates directly to the notion of Bayesian Surprise as we take interest in the divergence between the PMI distribution of each term between our expectation space and our new observation \citep{baldi2010bits, grace2017encouraging}. Therefore, we create $W_{TNT} = {w_1, w_2, ..., w_m}$ the intersection set between $W_T$ and $W_{NT}$ and composed of $K$ words. For each word $w_k$ in $W_{TNT}$, we can estimate the divergent surprise as:

\begin{equation}
    Divergent Surprise = \frac{1}{K} \sum_{i=1}^{K} D_{JS}(PPMI_{w_i}||QPMI_{w_i}) 
\end{equation}

While the first metric of \textit{Cultural New Surprise} represents a violation of the expectation space as an extension or redefinition, the second notion of \textit{Cultural Divergent Surprise} represents how the modification of the ensemble of combination distribution violates the general view of our cultural expectation of a product.

\section{Experimental Design}%
%\label{sec:exp}%

In this section, we introduce how we created \textit{GlobalFusion}, our dataset on cultural variations in cooking recipes. We additionally present the experimental choices and the cultural distance measures employed to evaluate and compare our proposed cultural novelty framework.

\subsection{Dataset}
%\label{sec:data}%

\begin{figure}
    \centering
        \centering
        \includegraphics[width=0.75\textwidth, trim={40mm 30mm 40mm 25mm}]{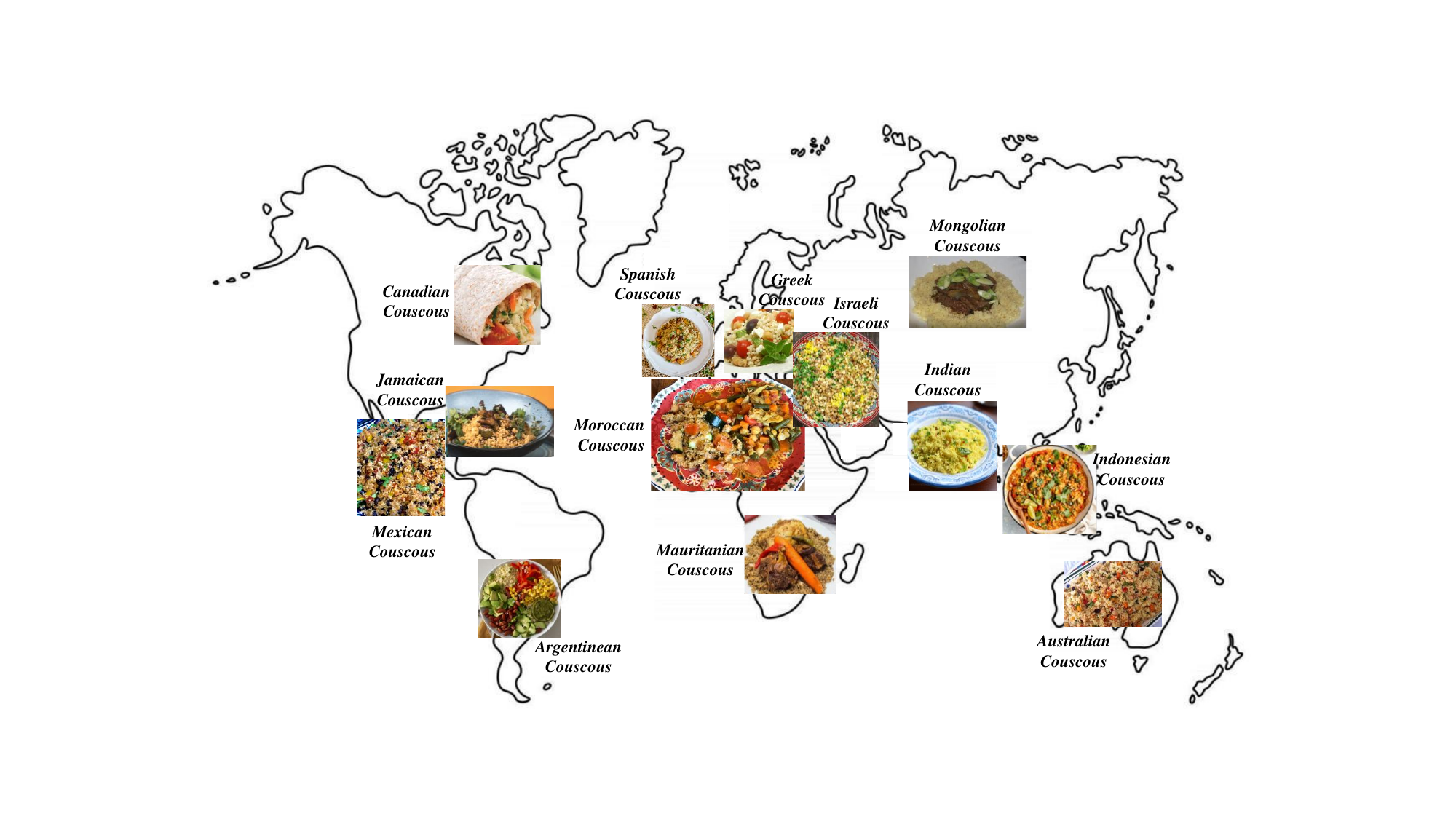}
        \caption{Illustration of Moroccan couscous recipes within our dataset \textit{GlobalFusion}. The variations are from diverse  countries such as Mauritanian couscous (n’gommou), One Pan Creamy Thai-Inspired Peanut Chicken Couscous, or Jamaican Jerk Couscous}
        \label{fig:world_map}
        %\Description{Examples on some cultural variations in GlobalFusion}
\end{figure}

To build our dataset on cultural novelty, we used RecipeNLG \citep{bien-etal-2020-recipenlg}. The dataset notably comprises the title, a recipe description, and a list of ingredients of more than 2M recipes. To extract our culturally aware knowledge bases and variations, we kept the recipes where we could identify a country in the title. We ended up with 135 385 recipes from 173 different countries. Then, we defined product names as typical dishes, for example, "Lasagna," "Couscous," or "Beef Curry," which were associated with a country as our cultural variable. We have manually selected the dishes based on several conditions: (1) the dish must include enough recipe variation for evaluation; (2) the dishes should present geographical and ingredient diversity; (3) the dish must have enough ingredients, should be a dish on its own, and should not follow normative constraints. Using the recipe title to match the proper dishes, we extracted each cultural product's associated recipe description to establish our knowledge space. We also identified variations of these dishes as recipe descriptions from different countries. Figure \ref{fig:world_map} presents an illustration of an image of recipes extracted with this approach for the couscous recipes, where the knowledge space is composed of 167 Moroccan recipes and 426 cultural variations coming from 35 different countries of all regions of the world.

We also disclose two couscous recipes, one from Morocco and one from Thailand, to provide an example of recipe preparation and the type of semantic variation we will study with our metrics.

\begin{table*}[t]
\centering
\begin{tabular}{p{0.45\linewidth} p{0.45\linewidth}}
\textbf{Moroccan Couscous} & \textbf{Jamaican Chicken Couscous} \\
\hline
Sprinkle the lamb with Moroccan spice. Heat the oil in a pan, add the lamb fillets and cook until the lamb is cooked as desired. 
    Remove the lamb to a warm serving plate.
    Add the onion, garlic, red pepper and carrot to the pan and cook for 2-3 minutes.
    Add the tomatoes and juice, and the Baby Peas, stirring to combine. Cover and cook over a low heat for 10 minutes.
    Spoon the sauce onto the prepared couscous. Cut the lamb into slices and serve on top of the sauce. & In a large saucepan, combine pineapple and 1/2 teaspoon salt; bring to a boil. Stir in couscous. Remove from heat; let stand, covered, 5 minutes or until liquid is absorbed. Fluff with a fork.
Meanwhile, in a shallow bowl, mix flour and 2 tablespoons cilantro. Toss chicken with jerk seasoning and remaining salt. Add to flour mixture, a few pieces at a time, and toss to coat lightly; shake off any excess.
In a large skillet, heat 1 tablespoon oil. Add a third of the chicken; cook 1-2 minutes on each side or until no longer pink. Repeat twice with remaining oil and chicken. Serve with couscous. If desired, sprinkle with cilantro.
 \\
\end{tabular}
\caption{Two culturally distinct couscous recipes from our dataset, illustrating the adaptation of couscous across cultural contexts.}
\label{tab:recipe_examples}
\end{table*}

\begin{figure}[!htbp]
    \centering
        \centering
        \includegraphics[width=0.95\linewidth]{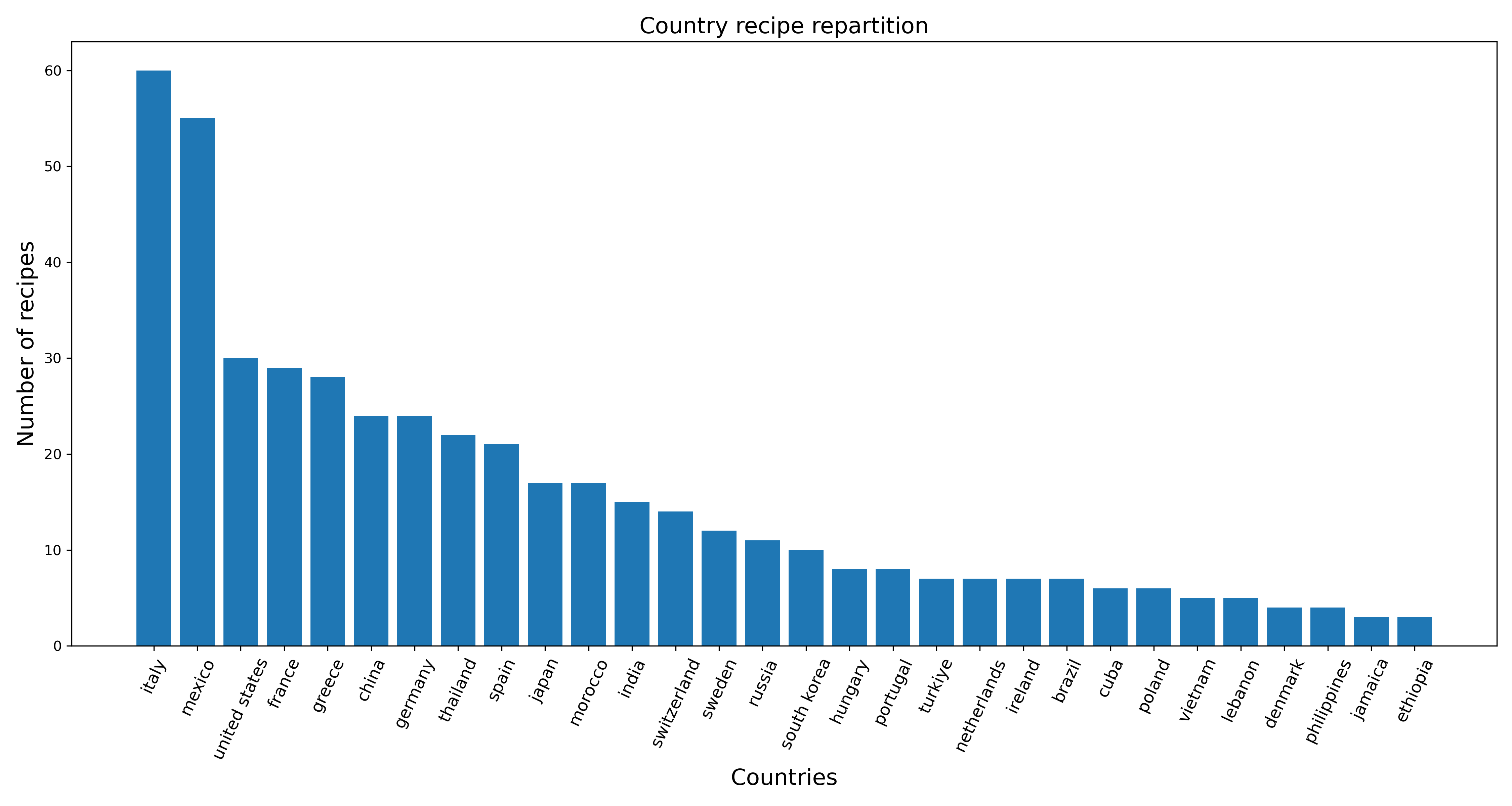}
        \caption{Country repartition of recipes in GlobalFusion.}
        \label{fig:country_rep}
\end{figure}

We identified 500 dishes, associated with 54 various countries from different geographical regions, forming our pairs of cultural products (product, country). Figure \ref{fig:country_rep} presents the detailed repartition of countries of our selected dishes.

We used the title to match a recipe and extracted the description to create our knowledge base. Each dish's knowledge space comprises an average of 60.98 recipe descriptions and is paired with 208.96 recipe descriptions of new variants coming from 19.28 different countries, for a total of 104,710 recipes variations. The specific protocol for data collection, ensuring respect for several conditions presented and detailed statistics on the dataset are provided in the appendix \ref{sec:annex_data}.

\subsection{Experiment and Evaluation}
\label{sec:eval}

\paragraph{Encoding:} 
Since we evaluate semantic divergence, we retain only nouns, adjectives, adverbs, numbers, and verbs, as other grammatical categories do not provide semantic information. Additionally, we lemmatize the text to reduce morphosyntactic variations. To measure the divergences between two recipes, we modeled the documents based on their word distribution. For the surprise metric, we estimate the PMI with a standard sliding window of size 3. We provide a more detail view of all text processing in the appendix \ref{sec:annex_results}.

\paragraph{Control variable:}
We have created several variables that we have used to control the analyses of our results or as mediation variables when measuring the marginal contributions of our metrics. We first have created a variable for \textit{Lexical diversity}. This variable represents the ratio between the number of unique tokens compared to the total number of words in the variation recipe. Then we have the variable \textit{New ingredients}, which measures the ratio of new ingredients compared to the total number of ingredients in the variation recipe. Finally we have the variable \textit{Length ratio}, which control the length of the description of the variation recipe with the average length of a text in our knowledge space.

\paragraph{Cultural distances:}
To establish that our framework captures cultural novelty, we compare our metrics with several distances highlighting different cultural aspects. 

\begin{itemize}
\item We use the Inglehart–Welzel Cultural Map \cite{WVSCultMap} provided by the World Value Survey. This global cultural map scores society on two dimensions: (1) Traditional values versus Secular-rational values and (2) Survival values versus Self-expression values. Following the work in \cite{tao2024cultural}, this two-dimensional measurement of culture allows us to estimate the Euclidean distance between countries studied.
\item We use the linguistic distance dataset\footnote{\url{http://dow.net.au/?page_id=32}}. We used the data collected from 2015 which followed the research protocol from \cite{gordon2005ethnologue}. This distance represents the distance between the two closest major languages for each pair of countries. 
\item We again use the religious distance dataset\footnote{\url{http://dow.net.au/?page_id=35}} collected in 2015 to measure the religious distance between countries. The religious distance measures the distance between the two major religions in a pair of countries. 
\item Finally, we compute a geographical distance in kilometers between countries based on each country's capital's latitude and longitude coordinates.
\end{itemize}

The dataset and the code are available on our Github page\footnote{\url{https://github.com/fcarichon/Cultural_Novelty}}.

\section{Results}
\label{sec:results}

\begin{figure*}
    \centering
    \begin{minipage}[b]{0.284\textwidth}
        \centering
        \includegraphics[width=\textwidth]{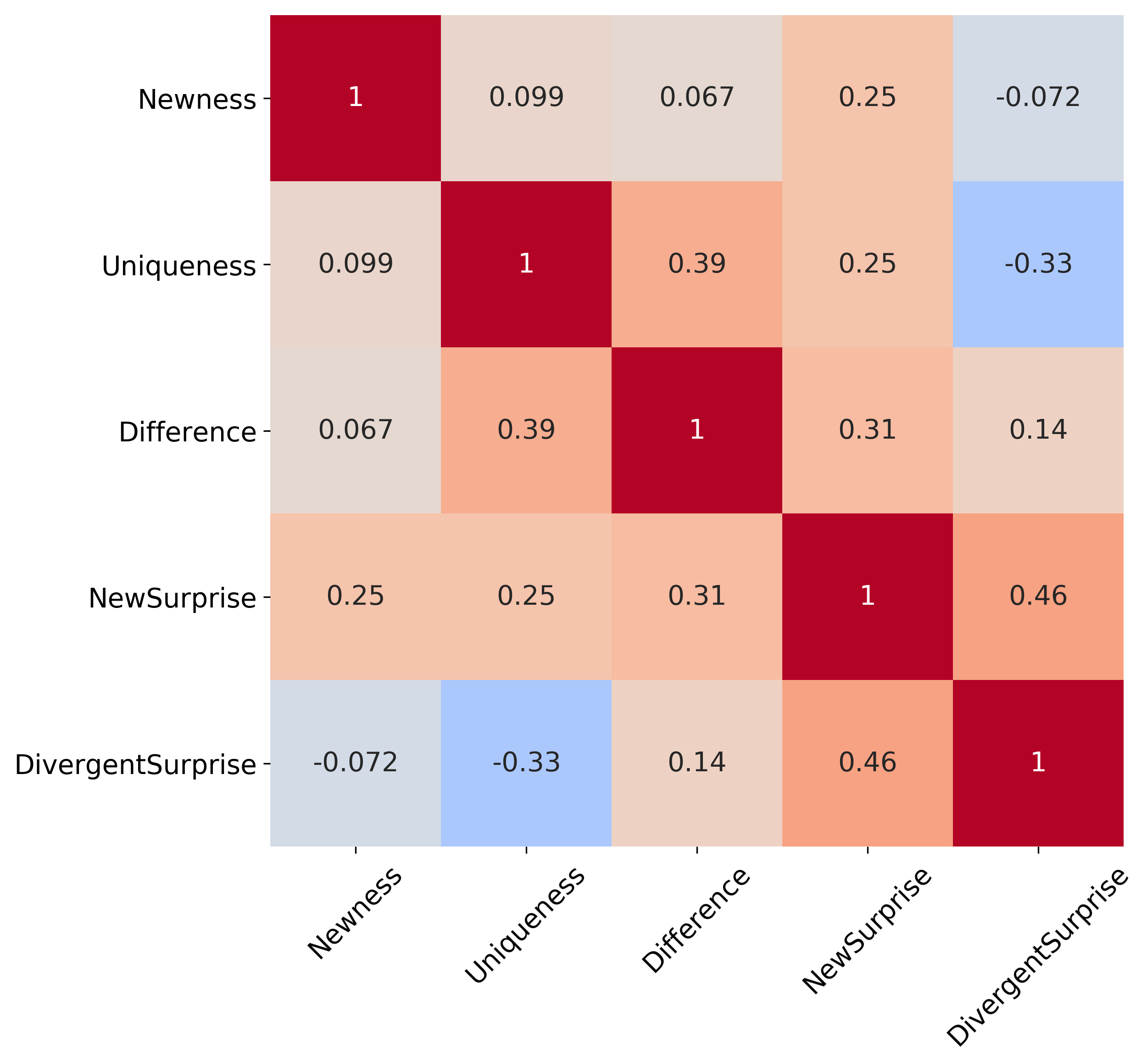}
        \subcaption{Pearson Correlation}
        %\label{fig:fig1}
    \end{minipage}
    \hfill
    \begin{minipage}[b]{0.22\textwidth}
        \centering
        \includegraphics[width=\textwidth]{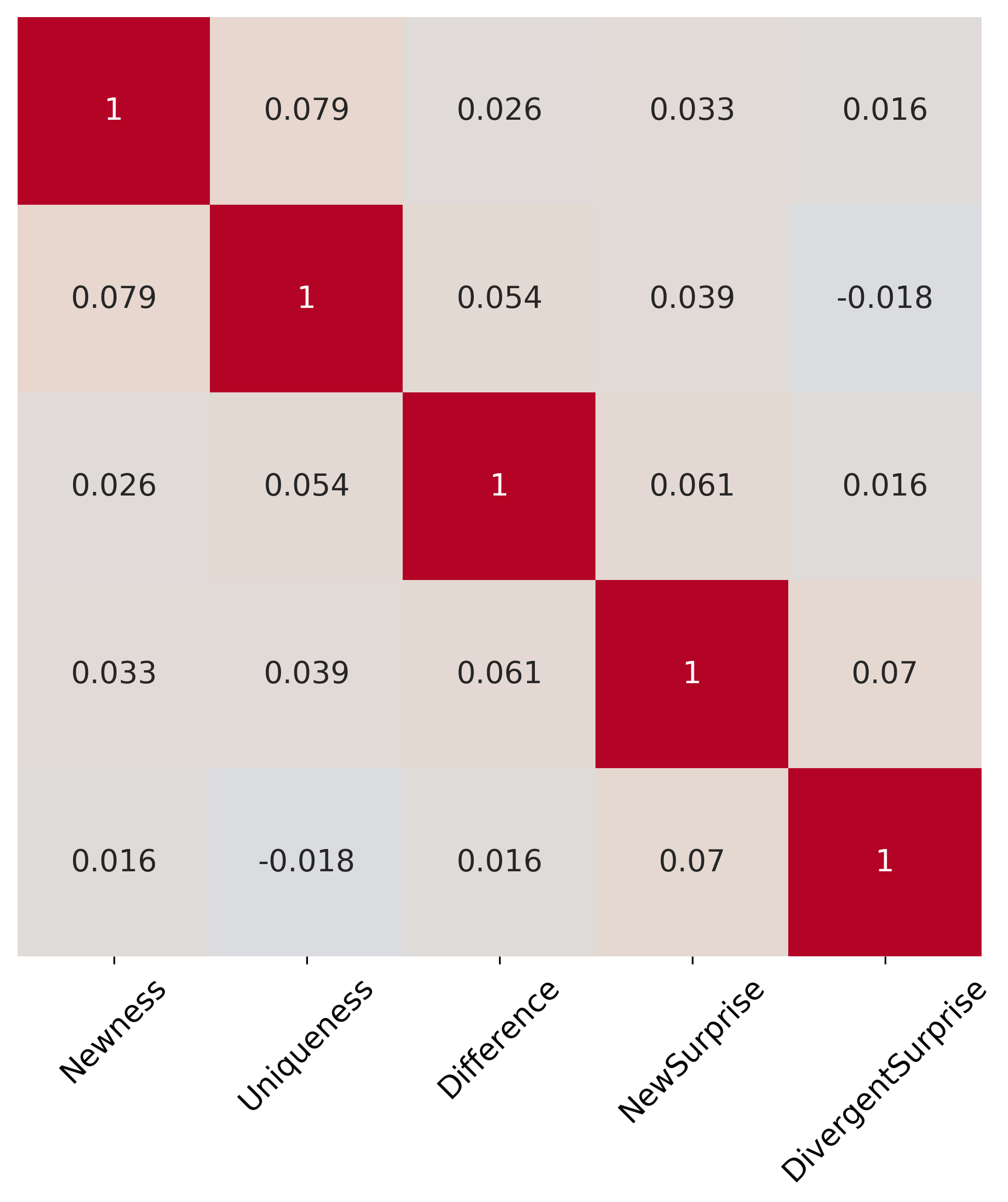}
        \subcaption{Kendall Tau Correlation}
        %\label{fig:fig2}
    \end{minipage}
    \hfill
    \begin{minipage}[b]{0.252\textwidth}
        \centering
        \includegraphics[width=\textwidth]{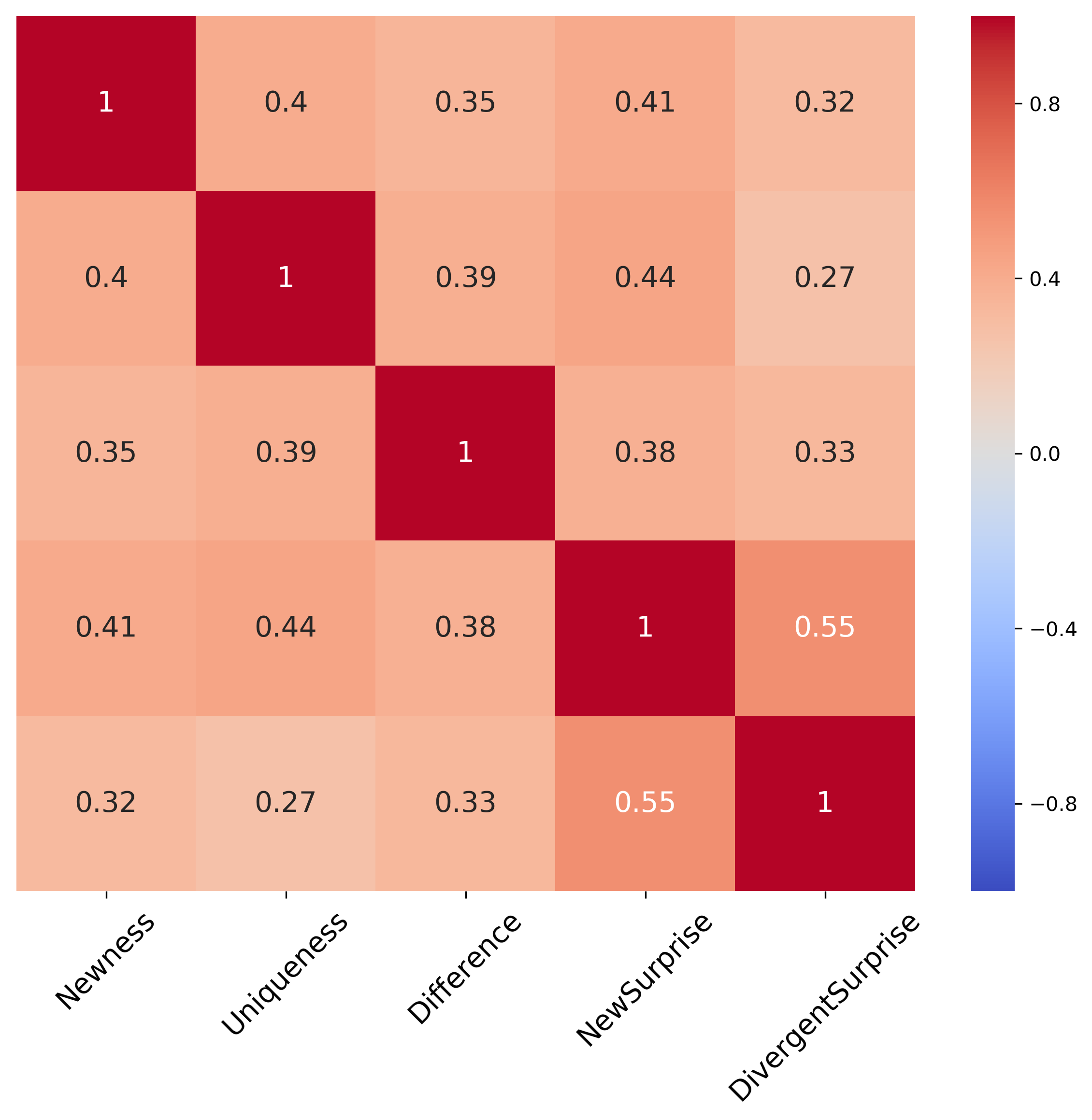}
        \subcaption{Ranked Bias Overlap}
        %\label{fig:fig2}
    \end{minipage}
    \caption{Correlation coefficients between our 5 different variables for novelty.}
    \label{fig:Correaltions}
\end{figure*}

While relying on information-theory-based techniques has shown their utility for different novelty tasks, it remains essential to demonstrate first that the metrics can be mobilized together to establish different and complementary outcomes. Therefore, we first measure the Pearson correlation between our metrics. We have also ranked the recipes according to their score on each metric and calculated the Kendall rank correlation coefficient and the rank-biased overlap (RBO) between ranked list. Figure \ref{fig:Correaltions} displays the strength of the correlations between our variables for these three types of analyses. Table \ref{tab:Correaltions} in the appendix \ref{sec:annex_results} presents the detailed results with the coefficients and their associated p-values, presenting the statistical significance of these correlations.

Most of our variables generally have a Pearson correlation value close to 0.3 with the other variables, indicating a strong relationship between them. It is expected since all these metrics encompass a notion of novelty or a divergence toward a known distribution. The first exception is \textit{Newness}, which correlates weakly with our other variables except \textit{New Surprise}, though it would make sense since both metrics capture the apparition of new terms. More surprisingly, we observe the strong negative correlation between \textit{Uniqueness} and \textit{Divergent Surprise}, meaning that shifting from the knowledge space to the expectation space modifies the measure of novelty to the prototypical distribution. Moreover, by observing the Kendall correlation and the RBO metrics we can affirm that each metric emphasizes different facets of information since there is no significant link between the ranking of the documents and that the overlap between these rankings is relatively low. 

To point out the disparities between our variables even more, we measure their correlation with some metadata available with our dataset. The results of these analyses are included in table \ref{tab:correl_meta} in the appendix \ref{sec:annex_results}. We can notably observed the notable correlation between the quantity of new ingredients and the notions of \textit{Surprise}, especially for \textit{New Surprise}. We can also observe a difference between \textit{Newness}, \textit{Uniqueness}, and \textit{Difference} and the notions of \textit{Surprise} since the latter correlate positively with the total number of ingredients in the recipe. This highlights the contrast in the representations of the knowledge and expectation spaces, capturing these variations in novelty. Both \textit{Newness} and \textit{New Surprise} correlate negatively with the number of documents in the knowledge base, which underlines a notion of familiarity with the product in these concepts, since the less knowledge of a recipe, the easier it is to bring novelty with term appearance. Finally, we can once again observe the Difference between \textit{Uniqueness} and \textit{Divergence Surprise} when it comes to lexical diversity, since the more diverse or rich the text is, the more it tends to the average corpus representation penalizing \textit{Uniqueness} while it allows to accentuate or create unobserved term collocations. These different analyses allow us to conclude confidently that these five metrics highlight different phenomena, thus making them relevant in measuring different aspects of novelty. 

\begin{figure}
    \centering 
    \includegraphics[width=\linewidth]{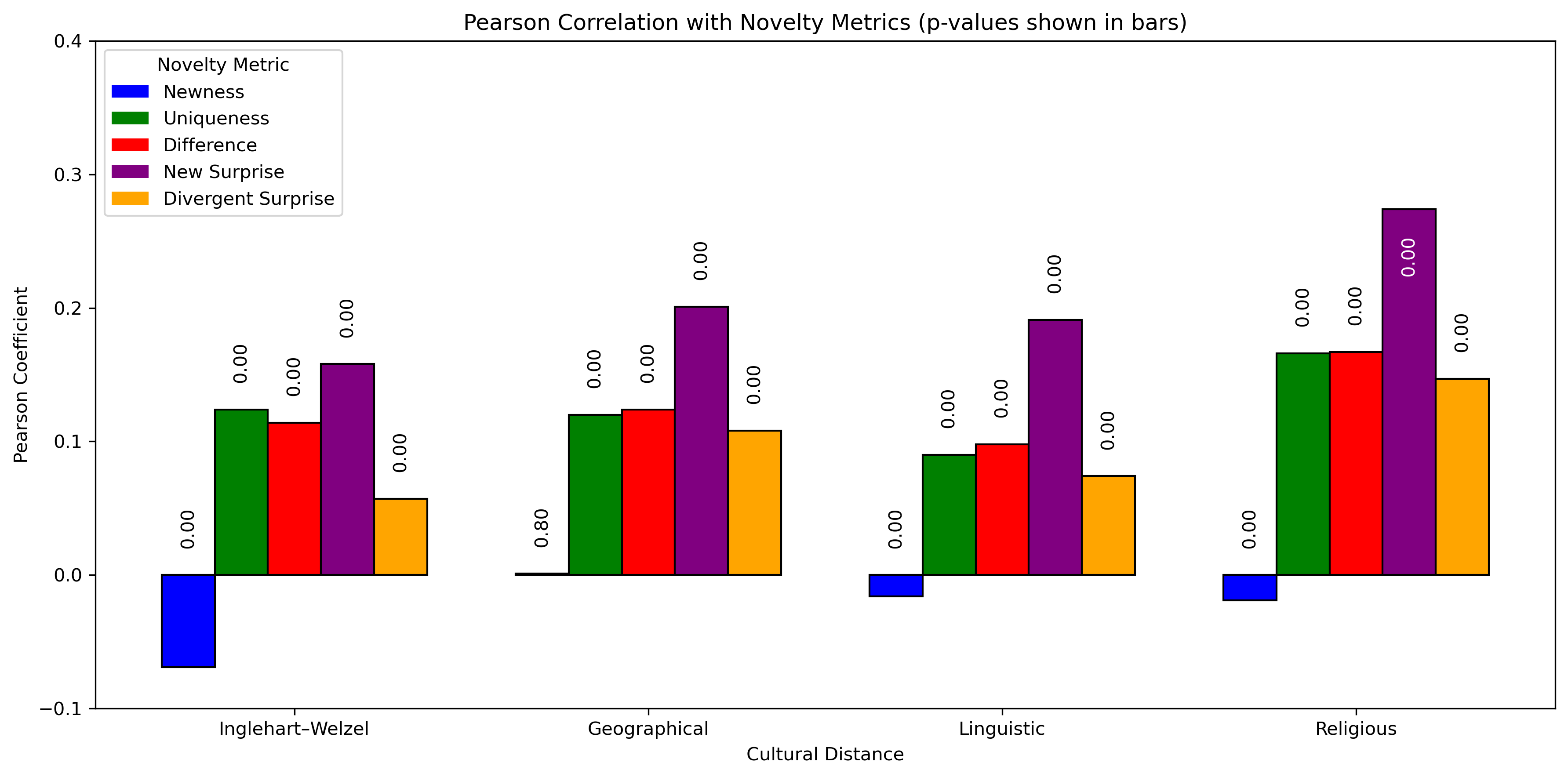}
    \caption{Correlations between our 5 novelty metrics and 4 distinct cultural distances. The bars' height highlights the strength of the coefficient, while the p-value is indicated above each bar.}
    \label{fig:NovelCultCorrel}
\end{figure}

We can now study the relationship between our metrics and different dimensions of cultural differences. Figure \ref{fig:NovelCultCorrel} presents the correlation between our metrics and cultural distances introduced in section \ref{sec:eval}. Table \ref{tab:NovelCultCorrel} in appendix \ref{sec:annex_results} presents the detailed results of these analyses.

\begin{figure*}
    \centering 
    \includegraphics[width=0.75\textwidth]{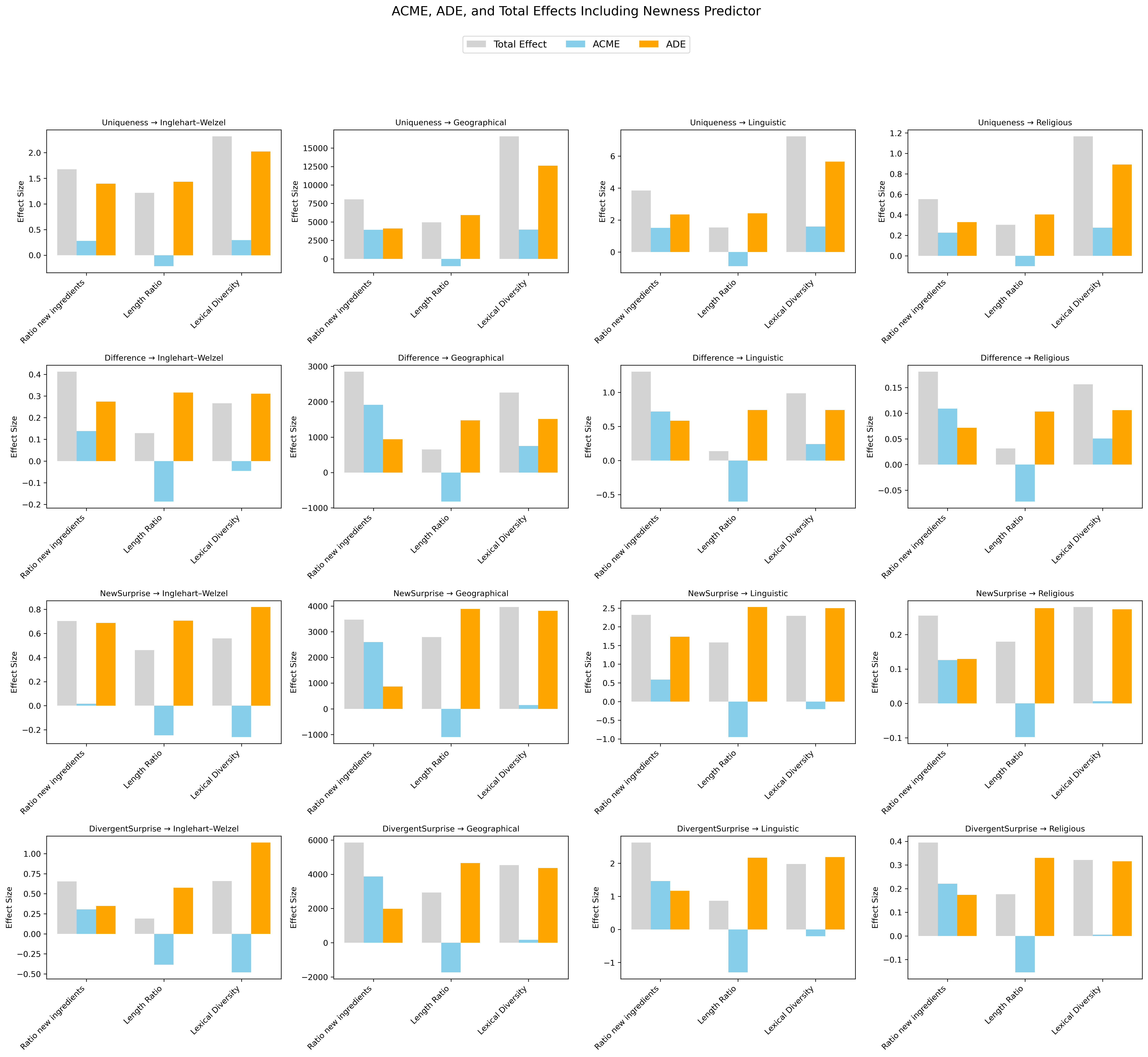}
    \caption{Mediation effects of the lexical diversity, the ratio length, and the ratio of new ingredients. The graphic includes the total effect, the average causal mediation effect (ACME), and the average direct effect (ADE).}
    \label{fig:mediation}
\end{figure*}

To solidify and extend these preliminary results, we completed our analyses by performing Ordinary Least Square (OLS) regressions, considering each cultural distance as a dependent variable. We have also considered adding the control variables introduced in section \ref{sec:eval}, and the table \ref{tab:GenRegression} in the appendix \ref{sec:annex_results} presents our models' results concisely. The Pearson's correlation and the regression analyses tend to confirm that novelty metrics are relevant proxies for measuring cultural distance. To better grasp the contribution of our novelty metrics to the different cultural distances, we also analyzed the marginal contribution of each variable. The detailed results are provided in table \ref{tab:marginal} in the appendix \ref{sec:annex_results}. We observed that their contribution varied significantly from their contribution in the entire model. This suggests mediating effects from our control variables on our metrics. Therefore, we have examined the marginal contribution of relevant novelty variables mediated by our control variables. Figure \ref{fig:mediation} presents the results of the mediation analyses for all variables except \textit{Newness} because the effects are often non-significant and out-of-scale for a clear visualization. The detailed results of all mediation analyses are provided in the table \ref{tab:marginal_med} in the appendix \ref{sec:annex_results}.

The first obvious observation that we can make is that all our metrics contribute to measuring some aspects of cultural distances. When considering mediation's effects, we can see the effect of new ingredients of both \textit{Surprise} metrics reducing their direct effect on all cultural metrics. The mediation effect of lexical diversity indicates an increasing direct effect of \textit{Uniqueness} on cultural distances. Other mediation effects seems marginal. Acknowledging these relations, we can depict a portrait of the relationship between cultural distances and novelty.
\begin{itemize}
\item \textbf{Inglehart–Welzel Cultural distance:} The distance remains strongly correlated with \textit{New Surprise} despite the mediation effect. The second variable having the most impact is \textit{Uniqueness}, and  the variable has the strongest effect when measuring its direct effect on this cultural distance. Then \textit{Difference} still has a strong effect on the variable, then comes \textit{Newness} and \textit{Divergent Surprise} whose contributions are more marginal but still statistically significant, especially when controlling by the ratio of new ingredients. Meaning that all dimensions of novelty are relevant to reflect the Inglehart–Welzel Cultural distance from textual analysis.
\item \textbf{Geographical distance:} While the \textit{New Surprise} effect seems strong if we look at the marginal contribution, its effect is almost entirely included by the ratio of new ingredients and its direct effect on the geographical distance between countries becomes difficult to interpret, making it irrelevant for this type of analysis. On the contrary, \textit{Uniqueness} is a significant variable, especially when we measure its direct effect accounting for lexical diversity. Then \textit{Difference} and \textit{Divergent Surprise} also strongly affect this distance. Finally, \textit{Newness} is not a good indicator of geographical distance.
\item \textbf{Linguistic distance:} While considering mediation effects, we can safely consider \textit{New Surprise} and \textit{Uniqueness} as the two variables with the most substantial effects on linguistic distance. However, with the control variables included in the model, it is interesting to note that the \textit{Uniqueness} effect becomes negative, indicating the relation with other variables can dramatically impact its interpretation. The \textit{Difference} also presents a significant effect on the linguistic distance. For the \textit{Divergent Surprise}, the individual effect of the variable is more complex to interpret because most of the effect of the variable is mediated by the ratio of new ingredients, making it less relevant to consider here. Despite being statistically significant, the contribution of \textit{Newness} does not seem to explain linguistic distance and its effects become non-significative when mediated by the recipe's length.
\item \textbf{Religious distance:} This distance is the best explained by our general model displaying a strong link with novelty. Once again, \textit{Uniqueness} displays the most substantial effect on this distance, followed by the \textit{Difference}, highlighting the importance of these two metrics in measuring cultural distance. Despite being a third good variable for this distance, when considering the mediation effect of new ingredients, the contribution of \textit{New Surprise} become less important. \textit{Divergent Surprise} also shows a significant positive effect on religious distance and, despite its weaker tie, continues to be an interesting variable to consider. Finally, the marginal contribution of \textit{Newness} does not seem relevant and its effects, again, depend on the length of the recipe. 
\end{itemize}

\section{Discussion}
\label{sec:discuss}
The present study took an interest in the link between textual novelty detection, defined as the detection of some new information that is offered regarding previous knowledge \cite{ghosal-etal-2022-novelty}, and cultural novelty as the subjective perception of cultural differences \cite{wilczewski2023cultural}. To establish this relationship, we first proposed a new dataset \textit{GlobalFusion} composed of cooking recipes linked to a specific country and variations adapted by different communities. Building a relevant knowledge space based on products whose textual description represents proxies for cultural aspects allowed us to build metrics that make it possible to interpret textual divergences as an indicator of cultural novelty and, thus, cultural differences. This framework exhibits the possibility of paralleling these two notions and shows the potential of natural language processing methods to represent this complex notion. In particular, we show that well-established general novelty metrics based on distribution divergence are efficient tools to measure cultural distance between the description of cultural products. Despite having relatively low R-square scores, we still observe significant and positive correlations that prove the utility of novelty detection techniques in language at least captures a small portion of such cultural differences. The gap is also easily understandable since these metrics rely on the partial descriptions of cultural products that do not embed the full spectrum of what constitutes cultural distances. However, this study demonstrates that a link exists and that some dimensions of language can be used to give a specific view of cultural differences. 

Another interesting point that increases the relevance of our metrics to capture novelty in such cultural products is their relation with crucial recipe aspects that also embed the notion of cultural novelty. In fact, innovation in food processing is perceived as a modification of ingredients, food preparation, its shape, its taste, or its packaging \cite{guerrero2009consumer}. The link emphasized between \textit{Surprise} and, in a smaller portion, \textit{Difference} metrics to the ratio of new ingredients and the \textit{Uniqueness} to recipe lexical diversity describe some of these properties. It also highlights part of the limitation of our model, which is to capture the entirety of perceived innovation, since the textual information in our dataset cannot capture notions of shape, size, and taste. 

We also discuss the case of text representation and divergence metrics as tools for measuring novelty. We deliberately chose not to enhance text representation through methods like topic modeling or contextual embeddings derived from recent LLMs. While we acknowledge that such techniques could improve novelty modeling, they would complicate the interpretation of our results. We also recognize the limitation of divergences-based approaches for novelty detection since such divergences could stem from complex phenomena beyond novelty. However, our choices become evident in an evaluation context, such as assessing the quality of LLM-generated content. Suppose distribution divergence techniques are to be used as metrics to evaluate LLMs' capacity for generating culturally aware novelty. In that case, it is crucial to ensure that the target of evaluation is not embedded within the measurement system itself \cite{Strathern1997}. Using advanced text encoding approaches could introduce confounding factors, making it difficult to determine whether correlations with our cultural distance metrics arise from the language data or the encoding model. Finally, as LLMs training and fine-tuning rely on entropy-based learning, similar to the divergence measures employed in this study, a fine-grained understanding of how language distributions evolve across cultural contexts could offer valuable insights. Such research could inform strategies to fine-tune LLMs, reducing their inherent bias toward the dominant Anglophone culture \cite{tao2024cultural}.

As the field of AI continues to expand, incorporating theoretical frameworks from social sciences becomes increasingly essential to address the ethical and cultural implications of AI tools. This aligns with recent calls in the literature to integrate sociological and management theories into AI research, especially for LLMs, to design systems and metrics better aligned with human values and societal needs \cite{liu2023trustworthy}. This study offers a concrete contribution in that direction by bridging the gap between artificial intelligence research and theoretical frameworks from sociology and management. By applying several transformations of the Jensen-Shannon divergence to measure the different kinds of novelty of the adaptation of cultural products in various countries and communities, we demonstrate how computational tools can integrate sociocultural insights to address complex questions about language, culture, and creativity. This interdisciplinary approach is critical for developing metrics that are not only robust but also contextually relevant in evaluating artificial intelligence systems, particularly as they aim to produce content that reflects diverse cultural perspectives.

\section{Limitations and Future research avenues}
\label{sec:limits}

A key limitation of our work is its reliance on a single dataset focused on one type of cultural product due to the lack of datasets dedicated to cultural novelty. This is compounded by our novelty metrics reflecting culture through the precise cultural representation in the knowledge space. However, developing \textit{GlobalFusion} involved a substantial manual effort to extract country-related recipes, balance representation and diversity, and ensure data sufficiency. It also required minimizing noise and cultural misappropriation in title matching and addressing biases from varying writing norms. While literature, news, and movie synopses are promising directions for future work, assigning a cultural identity to them is more complex than for recipes, where users explicitly self-identify the country of origin. For instance, defining a movie's cultural origin, whether by its producer, director, actors, or dominant themes, poses methodological challenges requiring careful consideration to avoid biases.

A second limitation stems from our selection strategy, which introduces several cultural representation biases. First, we struggled to include a complete diversity of countries from Africa, South America, or Oceania due to RecipeNLG's limited coverage, narrowing the scope of our analysis. Additionally, certain countries use appropriate cultural markers, such as countries' names, in their culinary traditions; for example, "Indian curry" is a British dish. Although we manually corrected identified cases, other misattributions may remain. Finally, our last limitation comes from the language used. All our recipes are sourced from anglophone websites, creating linguistic and cultural representation biases. Using anglophone data for foreign recipes introduces alteration preparation descriptions. However, multilingual recipe datasets are scarce, and translating recipes from original languages could introduce biases of modifying the phraseology of recipes or lacking grounding from these specific cultures \cite{hershcovich2022challenges, cao2024cultural}. 

This research opens avenues for exploring alternative divergence representations, such as multi-prototypical and conceptual language representations or entropy ranking-based approaches, to capture new dimensions of cultural novelty. Moreover, our metrics could be applied to estimate the propagation of novel recipes across geographical boundaries using temporal metadata. They also open avenues in the classic NLP field and LLM analyses. First, traditional measures like TF-IDF and cosine similarity may enhance these metrics; it could also help understand cultural representation in word embeddings. For LLMs, prompting models to generate either traditional Moroccan or Venezuelan couscous would allow for measuring how well they reflect culturally grounded adaptations. Finally, since LLM fine-tuning often leverages relative entropy \cite{ahmadian2024unsupervised}, our metrics align naturally with such adaptation tasks and could be integrated into model evaluation or mitigation processes.

\section{Conclusion}
\label{sec:ccl}

In conclusion, we have presented a comprehensive framework designed to measure cultural differences, including constructing a new dedicated dataset, \textit{GlobalFusion}. This framework also establishes the necessary conditions to define a data-based common ground, control novelty variations, and be able to highlight cultural differences in language. By leveraging novelty detection approaches, we have demonstrated significant correlations with various methods for measuring cultural distance, underscoring the potential of our framework in evaluating cultural differences in text production. We believe that this framework can be readily generalized to various contexts, tasks, and datasets commonly utilized within the NLP community. It paves the way for developing various LLM white-box evaluation techniques. We hope such techniques will enhance the understanding of cultural awareness in LLM systems and how their representations align with human behavior.
%Bibliography
\bibliographystyle{apalike}  
\bibliography{references}

\appendix 
\section{Dataset description}
\label{sec:annex_data}
In this section, we introduce further information on our dataset.
We created \textit{GlobalFusion} our dataset by extracting around 130K recipes from recipeNLG \cite{bien-etal-2020-recipenlg}, where we could identify the country or nationality in the title associated with the recipe. To extract the 500 dishes in our dataset, we followed a selection protocol based on the following conditions to identify the proper recipes for our analyses.

\begin{itemize}
\item \textbf{Recipe popularity/sample size:} Each dish selected should be associated with at least one main reference country to form the knowledge base. This base should be composed of a minimum of 2 different recipe texts and paired with at least 2 recipe variations whose divergence for this dish will be studied. If we take the example of the product being \textit{Lasagnas}, we have identified all countries associated with lasagna. If a country like \textit{Italy} respects that popularity limit for the knowledge space, we associate it with the product to constitute our cultural product \textit{Italian Lasagnas} composed of 109 recipes. Since we also have 257 recipes paired with \textit{Mexico}, we have also created the cultural product of \textit{Mexican Lasagna}.
\item \textbf{Recipe diversity:} We selected recipes from various country regions to ensure that our results are unrelated to a specific cultural perception type. To create these regions, we clustered countries based on the number of common ingredients in their recipes. Once clusters have been created, we selected several recipes from their most popular countries to respect our sample size constraint based on the overall recipe distribution per country. The results of the clustering as the recipe distribution graph per country are accessible respectively in appendix \ref{sec:annex_data}. 
\item \textbf{Recipe naming:} For each name, we have considered local variations of that name when extracting their associated recipes. For example, some recipes mentioned \textit{Crêpes: French Pancakes}. Therefore, when considering the cultural product composed by the pair (\textit{Pancakes}, \textit{France}), we made sure to add a pair (\textit{Crêpes}, \textit{France}) to match all relevant recipes in our dataset. We have also considered all local semantic variations of dish names. For the dish \textit{Pierogi}, our extraction strategy accounted for other writing, such as \textit{Piroshki}.
\item \textbf{Recipe name specialization:} We needed to ensure that each dish name matched the proper recipe and had enough but relevant variations. For cases such as \textit{Lasagnas} or \textit{Humus}, matching recipe titles is trivial because the name directly refers to the dish. However, for other cases, such as \textit{Curry} or \textit{Stew}, which are more generic, we needed to preserve names that allow for matching some cultural variations. Therefore, we have kept the most general names, guaranteeing we match the same dish: Not too generic as \textit{Curry} but not too specific as \textit{Red and Spicy Lentil Curry}. 
\item We have decided not to retain any recipe where the number of ingredients was too low on average, such as bread, to allow enough creativity and novelty to appear in the recipe descriptions.
\item We have discarded any dish that could be included in too many other dishes, such as sauces, because the variations would then not correspond to the original product studied, and thus, the novelty would not correspond to cultural variations.
\item We have discarded dishes that are restrained by naming norms such as cocktails, not allowing for variations because they would be directly renamed differently, thus not allowing us to create suitable matches. 
\end{itemize}

\begin{figure}
  \centering
 \includegraphics[width=\linewidth]{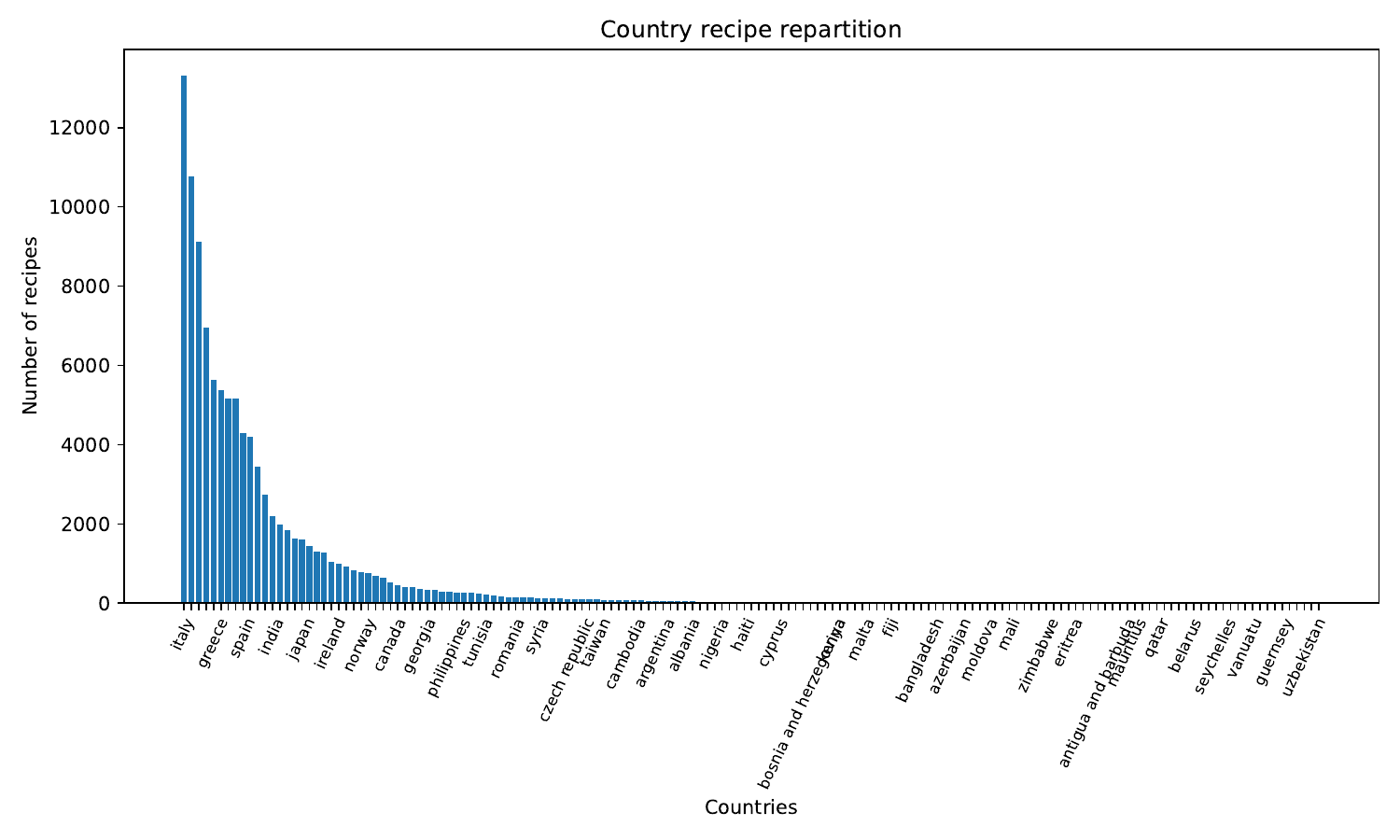}
  \caption{Total Country repartition in GlobalFusion.}
  \label{fig:countries}
\end{figure}

Figure \ref{fig:countries} introduces the total country repartition in the dataset. The countries most represented in our dataset are Mexico, Italy, Greece, the United States, and Germany. We then selected 500 recipes associated with a country to constitute our knowledge bases and to identify variations. The countries most represented in our 500 recipes are Mexico, Italy, Greece, the United States, and France.

\begin{figure}
  \centering
 \includegraphics[width=\linewidth]{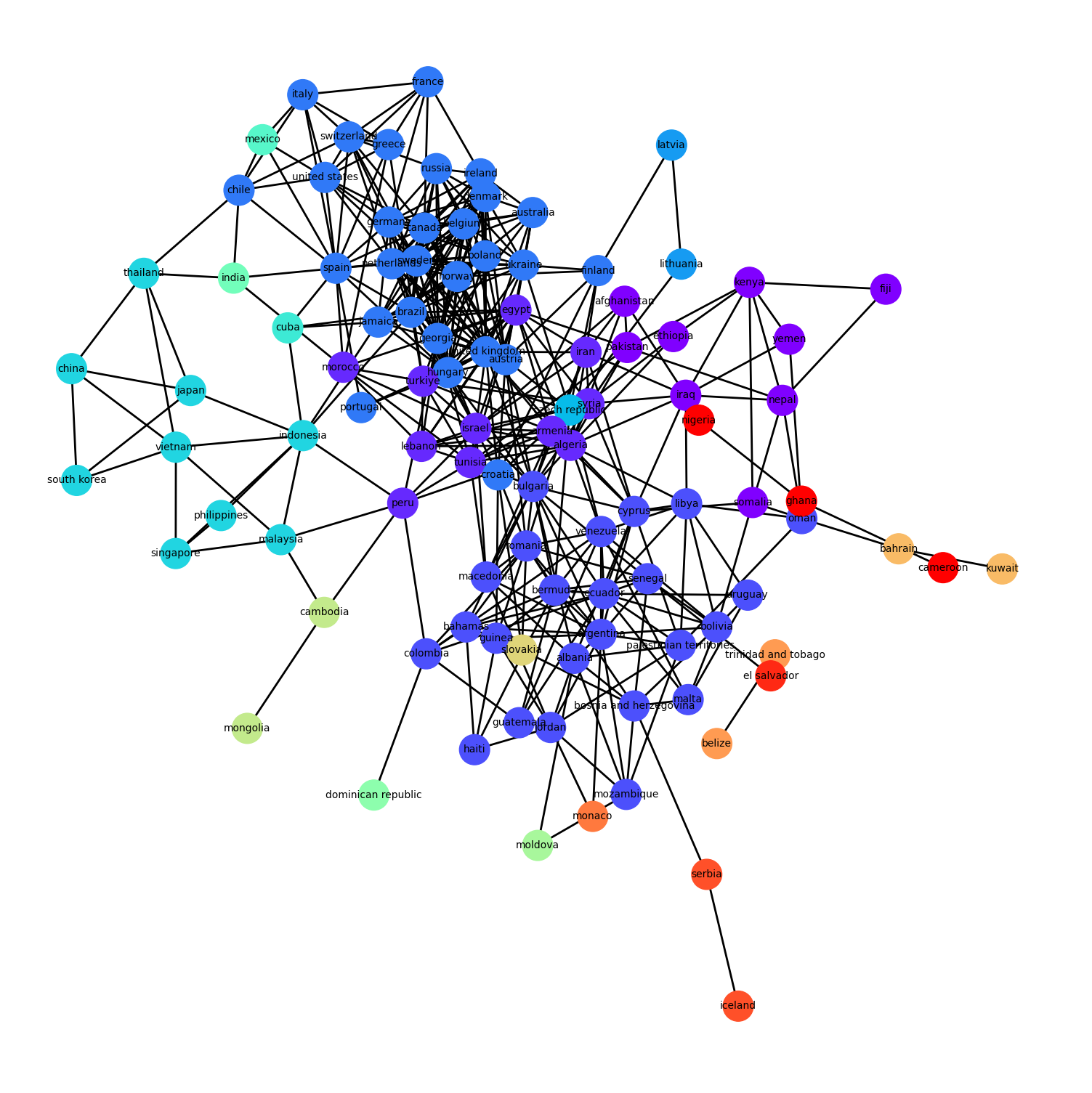}
  \caption{Country Clusters based on common popular ingredients.}
  \label{fig:clusters}
\end{figure}

To ensure a certain amount of diversity and representativeness in the country of origin of our 500 base recipes that still respect our popularity constraints, we have clustered countries based on their similarity of ingredients. More specifically, we created the set of their typical ingredients for each country by preserving their 20\% most occurring ones. Then, we measure the similarity between countries using a Jaccard similarity measure. We used the Newman Modularity clustering algorithm to obtain our groups. Figure \ref{fig:clusters} presents our clustering results. Once clusters were created, we ensured we had at least some recipes representing each country cluster in our 500 dishes. 

Once we have selected the 500 dishes we want to retain, we extract all recipes associated with these dishes. Recipes from the same country will constitute our knowledge base, and the ones from other countries will be considered variations for novelty evaluation. We also randomly excluded 30\% of the recipes from the origin country to include them in the variations to analyze variations from all origins. The table \ref{tab:data_stats} provides statistics about the recipe texts and other metadata. 

\begin{table*}
\caption{Statistics presenting the GlobalFusion Datasets. Numbers in parenthesis indicate standard deviation of the related indicator.}
\centering
  \label{tab:data_stats}
  \small
  \begin{tabular}{|l|c|c|c|c|}
    \hline
    \multicolumn{1}{|c|}{} & \multicolumn{1}{c|}{Knowledge} & \multicolumn{3}{c|}{Variations} \\
    \cline{3-5}
    \multicolumn{1}{|c|}{} & \multicolumn{1}{c|}{Base} & Same Country & Different Country & Total \\
    \hline
   % \hline
    Average number of Recipes & 60.98 (147.3) & 15.24 (37.26) & 193.75 (366.13) & 208.96 (293.95) \\
    \hline
    Average number of tokens per recipe & 543.75 (231.05) & 548.74 (246.67) & 572.64 (271.13) & 552.93 (261.26) \\
    \hline
    Average number of tokens after processing & 68.52 (28.12) & 68.86 (30.21) & 71.95 (33.07) & 69.77 (32.22)   \\
    \hline
    Number of ingredients & 136.3 (164.47) & 48.98 (58.8) & 315.92 (376.47) & 275.58 (329.85) \\
    \hline
    Number of new ingredients &  & 23.2 (28.15) & 289.77 (347.64) & 275.74 (346.22) \\
  \hline
\end{tabular}
\end{table*}

\section{Experiment details}
\label{sec:annex_results}

%%%%%%%%%%%%%%%%%%% TEXT PROCESSING %%%%%%%%%%%%%%%%%%%%%%%%%%%%%%%%%%
\subsection{Text processing and Encoding} 
Since our metrics perform at a word level, we have performed some text treatment to correctly measure semantic and cultural differences. We used the Spacy toolbox\footnote{https://spacy.io/} to perform all our treatments. Specifically, we first used their tokenizer for variable creation. We use the POS tagger to preserve only nouns, adjectives, adverbs, numbers, and verbs, since other grammatical categories will not provide semantic information. Then we employed Spacy's lemmatizer to reduce morphosyntax variations. Since we want to measure the divergence between $P_{prod_k;c_j} = P(w|T_{prod_k;c_j})$ and $Q_{prod_k} = P(w|NT_{prod_k;c_i})$, for each description of the recipe, we have counted the frequency of each word and divided it by the total number of words in that document to get a probability distribution of the recipe. When dealing with knowledge space corpus probability, we employ the same approach and count the number of times the word appears divided by the total number of words in the corpus. For the surprise metric, we estimate the PMI with a standard sliding window of size 3 based on the NLTK collocation tool\footnote{https://www.nltk.org/api/nltk.collocations.BigramCollocationFinder.html} for each recipe.

\subsection{Complementary results} 
%%%%%%%%%%%%%%%%%%%% CORREALTION BETWEEN METRICS %%%%%%%%%%%%%%%%%%%%%%%%

This table \ref{tab:Correaltions} shows the detailed analyses of the Pearson correlations between the results of the different metrics, as well as the Kendall Tau correlations both with their associated p-values, and the Ranked Bias overlap between the rankings of the recipes obtained with our metrics. The results show a significant correlation between our variables, highlighting their link to novelty. However, the recipes with the highest scores for each metric are not the same, and their rankings are not correlated. 

\begin{table*}
\caption{Correlation analyses between our 5 different variables for novelty.}
\centering
  \label{tab:Correaltions}
  \small
  \begin{tabular}{|l|c|c|c|}
    \hline
     & \makecell{Pearson Correlation\\(p-value)} & \makecell{Kendall rank correlation\\(p-value)} & \makecell{Rank-Biased Overlap} \\
    \hline
   % \hline
    Newness \& Uniqueness & $0.098$ ($9.21e-225$***) & $0.079$ $(0.493)$ & $0.4$ \\
    \hline
    Newness \& Difference & $0.067$ ($2.66e-105$***)  & $0.026$ ($0.528$) & $0.348$ \\
    \hline
    Newness \& New Surprise & $0.253$ ($0.0$***) & $0.033$ ($0.506$) & $0.414$  \\
    \hline
    Newness \&  Divergent Surprise & $-0.071$ ($3.3e-120$***) & $0.016$ ($0.516$) & $0.317$ \\
    \hline
    Uniqueness \&  Difference & $0.393$ ($0.0$***)  & $0.054$  ($0.508$) & $0.388$ \\
    \hline
    Uniqueness \& New Surprise & $0.25$ ($0.0$***) & $0.039$ ($0.512$) & $0.441$ \\
    \hline
    Uniqueness \& Divergent Surprise & $-0.325$ ($0.0$***) & $-0.018$ ($0.512$) & $0.271$  \\
    \hline
   Difference \& New Surprise & $0.311 $ ($0.0$***) & $0.061$ ($0.505$) & $0.376$ \\
    \hline
    Difference \& Divergent Surprise & $0.138$ ($0.0$***) & $0.016$ ($0.516$) & $0.33$ \\
    \hline
    New Surprise \& Divergent Surprise & $0.459$ ($0.0$***)  & $0.07$ ($0.442$) & $0.549$ \\
  \hline
\end{tabular}
\end{table*}

%%%%%%%%%%%%%%%%%%%%%%%%% CORREALTION METRICS / COOKING %%%%%%%%%%%%%%%%%%%%

We conducted complementary analyses to estimate the correlations between our divergence metrics and other variables related to cooking recipes, such as the number of ingredients and new ingredients, the number of recipes in the knowledge base, or the number of tokens in our recipes. Table \ref{tab:correl_meta} presents these results.

\begin{table*}[h]
\caption{Pearson correlation analyses between our 5 different variables for novelty and metadata associated to the recipes. The value in parenthesis indicate the p-value associated the correlation coefficient.}
\centering
  \label{tab:correl_meta}
  \small
  \begin{tabular}{|l|c|c|c|c|c|c|c|}
    \hline
     & \makecell{Number of \\ Ingredients} & \makecell{Ratio of new \\ ingredients} & \makecell{Knowledge \\ base size} & \makecell{Number \\ of tokens} & \makecell{Lexical \\ diversity} & \makecell{Ratio of token \\  number between the \\ KB vs Variations} \\
    \hline
   % \hline
     Newness & -0.135 (0.0***) & -0.294 (0.0***) & 0.177 (0.0***) & -0.087 (0.0***) & -0.147 (0.0***) & 0.16 (0.0***) \\
     Uniqueness & -0.355 (0.0***) & 0.138 (0.0***) & 0.167 (0.0***) & -0.524 (0.0***) & -0.554 (0.0***) & -0.275 (0.0***) \\
     Difference & -0.076 (0.0***) & 0.101 (0.0***) & 0.169 (0.0***) & -0.016 (0.0***) & -0.043 (0.0***) & 0.004 (0.184) \\
     New Surprise & 0.114 (0.0***) & -0.168 (0.0***) & 0.611 (0.0***) & 0.264 (0.0***) & 0.31 (0.0***) & 0.006 (0.05*) \\
     Divergent Surprise & 0.299 (0.0***) & 0.025 (0.349) & 0.226 (0.0***) & 0.56 (0.0***) & 0.615 (0.0***) & 0.027 (0.0***) \\
    \hline
\end{tabular}
\end{table*}

%%%%%%%%%%%%%%%%%%%% CORREALTION METRICS / CULTURE %%%%%%%%%%%%%%%%%%%%%%%%

Once we have validated that our metrics deliver various notions of novelty, we analyzed the Pearson correlations of these metrics with various notions of cultural distances. Table \ref{tab:NovelCultCorrel} presents these analyses. 

\begin{table*}
\caption{Pearson Correlation between our 5 novelty metrics and several cultural distances.}
\centering
  \label{tab:NovelCultCorrel}
  \small
  \begin{tabular}{|l|c|c|c|c|}
    \hline
     \makecell{Pearson Coefficient \\ (P-value)}& \makecell{Inglehart–Welzel  \\ Cultural Distance} & \makecell{Geographical Distance} & Linguistic Distance & Religious Distance \\
    \hline
   % \hline
    Newness & $-0.069$ ($0.0$***) & $0.001$ ($0.801$) & $-0.016$ ($0.0$***) & $-0.019$ ($0.0$***) \\
    \hline
    Uniqueness & $0.124$ ($0.0$***) & $0.12$ ($0.0$***) & $0.09$ ($0.0$***) & $0.166$ ($0.0$***) \\
    \hline
    Difference & $0.114$ ($0.0$***) & $0.124$ ($0.0$***) & $0.098$ ($0.0$***) & $0.167$ ($0.0$***) \\
    \hline
    New Surprise & $0.158$ ($0.0$***) & $0.201$ ($0.0$***) & $0.191$ ($0.0$***) & $0.274$ ($0.0$***) \\
    \hline
    Divergent Surprise & $0.057$ ($0.0$***) & $0.108$ ($0.0$***) & $0.074$ ($0.0$***) & $0.147$ ($0.0$***) \\
    \hline
\end{tabular}
\end{table*}

We have completed Pearson's correlation analyses with Ordinary Least Square Regression to measure the effect of our variables on cultural distance controlled by several linguistic aspects. Our complete model can be represented as:

\begin{equation}
    \begin{split}
	CulturalDistance = \beta_0 + \beta_1 * Newness + \beta_2 * Uniqueness + \\  \beta_3 * Difference + \beta_4 * Surprise + \beta_5 * NewIngredients \\ + \beta_6 * LexicalDiversity + \beta_7 * LengthRatio
    \end{split}
\end{equation}

Table \ref{tab:GenRegression} presents the detailed results of our model.

\begin{table*}
\caption{Results from our  Ordinary Least Square regressions with each cultural distance.}
\centering
  \label{tab:GenRegression}
  \small
  \begin{tabular}{|l|c|c|c|c|c|c|c|c|}
    \hline
    \multicolumn{1}{|c|}{} & \multicolumn{2}{c|}{\makecell{Inglehart–Welzel \\ Cultural Distance}} & \multicolumn{2}{c|}{Geographical Distance} & \multicolumn{2}{c|}{Linguistic Distance} & \multicolumn{2}{c|}{Religious Distance} \\
    \cline{2-9}
    \multicolumn{1}{|c|}{} & \multicolumn{2}{c|}{R-square: 0.054 (0.0***)} & \multicolumn{2}{c|}{R-square: 0.097 (0.0***)} & \multicolumn{2}{c|}{R-square: 0.06 (0.0***)} & \multicolumn{2}{c|}{R-square: 0.145 (0.0***)} \\
    \cline{2-9}
     \multicolumn{1}{|c|}{} & Coef. & $P>|t|$ & Coef. & $P>|t|$ & Coef. & $P>|t|$ & Coef. & $P>|t|$ \\
    \hline
   % \hline
    Const. & 1.23 & 0.0*** & 467.92 & 0.002** & 7.45 & 0.0*** & 0.25 & 0.0***\\
    \hline
    Newness & -16.64 & 0.0*** & -18220 & 0.0*** & -24.30 & 0.0*** & -1.74 & 0.0*** \\
    \hline
    Difference & 0.15 & 0.0*** & 548.46 & 0.0*** & 0.41 & 0.0*** & 0.04 & 0.0*** \\
    \hline
    Uniqueness & 0.002 & 0.973 & 5087.6 & 0.0*** & -0.33 & 0.06 & 0.34 & 0.0*** \\
    \hline
    New Surprise & 0.82 & 0.0*** & -470.16 & 0.0*** & 1.99 & 0.0*** & 0.05 & 0.0*** \\
    \hline
    Divergent Surprise & 0.19 & 0.0*** & 2875.8 & 0.0*** & 0.24 & 0.04* & 0.15 & 0.0*** \\
    \hline
    New Ingredients & 0.12 & 0.0*** & 4708.8 & 0.0*** & 1.15 & 0.0*** & 0.23 & 0.0*** \\
    \hline
    Ration Length & -0.11 & 0.0*** & -638.03 & 0.0*** & -0.79 & 0.0*** & -0.07 & 0.0*** \\
    \hline
    Lexical Diversity & -0.23 & 0.0*** & 222.27 & 0.0*** & -0.19 & 0.0*** & 0.02 & 0.0*** \\
  \hline
\end{tabular}
\end{table*}

%%%%%%%%%%%%%%%%%%%%%%%%%%%%%%%%%%% MARGINAL CONTRIBUTION TABLE %%%%%%%%%%%%%%%%%%

We also ran some complementary analyses to estimate the marginal contribution of each variable to the different cultural distances analyzed, compare them with the full model that includes the control variables, and better explain the observed Pearson correlations. The table \ref{tab:marginal} presents these results. We observe differences between these marginal contributions and the effect in the full model; knowing the correlation between our metrics and some metadata, such as the number of new ingredients, we also performed mediation analyses of this marginal contribution to have information on their direct effect on cultural distances. The table presents \ref{tab:marginal_med} these results. In this table, we do not present the mediation effects when the variable is already not significant in any way to the effect on cultural distances. 

\begin{table*}
\caption{Ordinary Least Square regressions with each cultural distance and individual novelty variables.}
\centering
  \label{tab:marginal}
  \small
  \begin{tabular}{|l|c|c|c|c|c|c|c|c|c|c|c|c|}
    \hline
    \multicolumn{1}{|c|}{} & \multicolumn{3}{c|}{\makecell{Inglehart–Welzel \\ Cultural Distance}} & \multicolumn{3}{c|}{\makecell{Geographical \\ Distance}} & \multicolumn{3}{c|}{\makecell{Linguistic \\ Distance}} & \multicolumn{3}{c|}{\makecell{Religious \\ Distance}} \\
    \cline{2-13}
     \multicolumn{1}{|c|}{} & Coef. & $P>|t|$ & $R^{2}$ & Coef. & $P>|t|$ & $R^{2}$ & Coef. & $P>|t|$ & $R^{2}$ & Coef. & $P>|t|$ & $R^{2}$ \\
    \hline
   % \hline
    Newness & -8.50 & 0.0 & 0.005 & 413.42 & 0.8 & 0.001 & -5.22 & 0.0 & 0.0 & -0.51 & 0.0 & 0 \\
    \hline
    Difference & 0.31 & 0.0 & 0.013 & 1472.58 & 0.0 & 0.016 & 0.72 & 0.0 & 0.01 & 0.10 & 0.0 & 0.028 \\
    \hline
    Uniqueness & 1.57 & 0.0 & 0.015 & 6571.88 & 0.0 & 0.014 & 0.0 & 3.01 & 0.008 & 0.47 & 0.0 & 0.027 \\
    \hline
    New Surprise & 0.70 & 0.0 & 0.025 & 3877.24 & 0.0 & 0.04 & 2.40 & 0.0 & 0.036 & 0.27& 0.0 & 0.022 \\
    \hline
    \makecell{Divergent \\ Surprise} & 0.55 & 0.0 & 0.003 & 4565.53 & 0.0 & 0.012 & 1.92 & 0.0 & 0.005 & 0.32 & 0.0 & 0.075 \\
  \hline
\end{tabular}
\end{table*}

%%%%%%%%%%%%%%%%%%%%%%%%%%%%%%%%%%% MEDIATION EFFECTS TABLE %%%%%%%%%%%%%%%%%%

\begin{table*}
\caption{Ordinary Least Square regressions accounting for the mediation effect of the ratio of new ingredient, the lexical diversity, and the ratio of token length between the reference and variations.}
\centering
  \label{tab:marginal_med}
  \small
  \begin{tabular}{|l|c|c|c|c|c|c|c|c|}
    \hline
    \multicolumn{1}{|c|}{} & \multicolumn{2}{c|}{\makecell{Inglehart–Welzel \\ Cultural Distance}} & \multicolumn{2}{c|}{\makecell{Geographical \\ Distance}} & \multicolumn{2}{c|}{\makecell{Linguistic \\ Distance}} & \multicolumn{2}{c|}{\makecell{Religious \\ Distance}} \\
    \cline{2-9}
    \multicolumn{1}{|c|}{} & \makecell{ACME \\ (p-value)} & \makecell{ADE \\ (p-value)} & \makecell{ACME \\ (p-value)} & \makecell{ADE \\ (p-value)} & \makecell{ACME \\ (p-value)} & \makecell{ADE \\ (p-value)} & \makecell{ACME \\ (p-value)} & \makecell{ADE \\ (p-value)}\\
    \hline
    \makecell{Newness \\ Ratio new ingredients} & \makecell{11.80 \\ (0.0***)} & \makecell{-11.03\\(0.0***)} & \makecell{127611\\(0.0***)} & \makecell{-26311\\(0.0***)} & \makecell{ 52.2\\(0.0***)} & \makecell{ -16.6\\(0.0***)} & \makecell{7.52 \\(0.0***)} & \makecell{-2.11\\(0.0***)}\\
    \hline
    \makecell{Newness \\ Lexical Diversity} & \makecell{-9.82\\(0.0***)} & \makecell{-7.29\\(0.0***)} & \makecell{-52020\\(0.0***)} & \makecell{6848.7\\(0.0***)} & \makecell{-36.5 \\(0.0***)} & \makecell{-0.94\\(0.37)} & \makecell{-4.52\\(0.0***)} & \makecell{0.04\\(0.65)} \\
    \hline
    \makecell{Newness  \\ Length Ratio} & \makecell{-4.52\\(0.0***)} & \makecell{-9.14\\(0.0***)}& \makecell{39650\\(0.0***)}& \makecell{6180.9\\(0.0***)}& \makecell{11.3 \\(0.0***)} & \makecell{-3.59\\(0.02**)} & \makecell{2.61\\(0.0***)} & \makecell{-0.14\\(0.09)} \\
    \hline
    \makecell{Uniqueness \\ Ratio new ingredients} &\makecell{0.27\\(0.0***)} &\makecell{1.39\\(0.0***)} & \makecell{3945.05\\(0.0***)}&\makecell{4113.64\\(0.0***)} & \makecell{1.50\\(0.0***)}&\makecell{2.34\\(0.0***)} & \makecell{0.22\\(0.0***)}& \makecell{0.32\\(0.0***)}\\
    \hline
    \makecell{Uniqueness \\ Lexical Diversity} & \makecell{0.29\\(0.0***)}&\makecell{2.02\\(0.0***)} & \makecell{3970.96\\(0.0***)}&\makecell{12613.23\\(0.0***)} & \makecell{1.59\\(0.0***)}&\makecell{5.65\\(0.0***)} &\makecell{0.27\\(0.0***)} & \makecell{0.89\\(0.0***)}\\
    \hline
    \makecell{Uniqueness \\ Length Ratio} &\makecell{-0.21\\(0.0***)} & \makecell{1.43\\(0.0***)}&\makecell{-994.41\\(0.0***)} &\makecell{5931.61\\(0.0***)} &\makecell{-0.89\\(0.0***)} &\makecell{2.42\\(0.0***)} &\makecell{-0.10\\(0.0***)} &\makecell{0.40\\(0.0***)} \\
    \hline
    \makecell{Difference \\ Ratio new ingredients} &\makecell{0.14\\(0.0***)} &\makecell{0.27\\(0.0***)} &\makecell{1911.91\\(0.0***)} &\makecell{936.77\\(0.0***)} & \makecell{0.71\\(0.0***)}&\makecell{0.58\\(0.0***)} & \makecell{0.11\\(0.0***)}&\makecell{0.07\\(0.0***)} \\
    \hline
    \makecell{Difference \\ Lexical Diversity} &\makecell{-0.04\\(0.0***)} &\makecell{0.311\\(0.0***)} &\makecell{749.95\\(0.0***)} & \makecell{1512.45\\(0.0***)}& \makecell{0.24\\(0.0***)}& \makecell{0.74\\(0.0***)}&\makecell{0.05\\(0.0***)} & \makecell{0.10\\(0.0***)}\\
    \hline
    \makecell{Difference \\ Length Ratio} & \makecell{-0.18\\(0.0***)}& \makecell{0.31\\(0.0***)}&\makecell{-821.99\\(0.0***)} &\makecell{1476.82\\(0.0***)} &\makecell{-0.60\\(0.0***)} &\makecell{0.74\\(0.0***)} &\makecell{-0.07\\(0.0***)} &\makecell{0.10\\(0.0***)} \\
    \hline
    \makecell{New Surprise \\ Ratio new ingredients} & \makecell{0.02\\(0.0***)}& \makecell{0.68\\(0.0***)}& \makecell{2596.15\\(0.0***)}&\makecell{872.82\\(0.0***)} &\makecell{0.58\\(0.0***)} & \makecell{1.73\\(0.0***)}& \makecell{0.12\\(0.0***)}& \makecell{0.13\\(0.0***)} \\
    \hline
    \makecell{New Surprise \\ Lexical Diversity} & \makecell{-0.26\\(0.0***)}& \makecell{0.82\\(0.0***)}& \makecell{142.62\\(0.0***)}& \makecell{3818.16\\(0.0***)}& \makecell{-0.20\\(0.0***)}& \makecell{2.5\\(0.0***)}&\makecell{0.006\\(0.0***)} &\makecell{0.27\\(0.0***)} \\
    \hline
    \makecell{New Surprise \\ Length Ratio} & \makecell{-0.24\\(0.0***)}& \makecell{0.71\\(0.0***)}& \makecell{-1096.56\\(0.0***)}&\makecell{3888.7\\(0.0***)} &\makecell{-0.95\\(0.0***)} &\makecell{2.52\\(0.0***)} &\makecell{-0.09\\(0.0***)} &\makecell{0.27\\(0.0***)} \\
    \hline
    \makecell{Divergent Surprise \\ Ratio new ingredients} &\makecell{0.30\\(0.0***)}& \makecell{0.35\\(0.0***)}& \makecell{3869.18\\(0.0***)}& \makecell{1985.39\\(0.0***)}&\makecell{1.46\\(0.0***)} & \makecell{1.17\\(0.0***)}&\makecell{0.22\\(0.0***)} &\makecell{0.17\\(0.0***)} \\
    \hline
    \makecell{Divergent Surprise \\ Lexical Diversity} & \makecell{-0.48\\(0.0***)}& \makecell{1.14\\(0.0***)}&\makecell{166.46\\0.04} &\makecell{4361.76\\(0.0***)} &\makecell{-0.20\\(0.0***)} &\makecell{2.18\\(0.0***)} &\makecell{0.005\\0.18} & \makecell{0.31\\(0.0***)} \\
    \hline
    \makecell{Divergent Surprise \\ Length Ratio} &\makecell{-0.38\\(0.0***)} & \makecell{0.57\\(0.0***)}&\makecell{-1730.2\\(0.0***)} & \makecell{4659.65\\(0.0***)}&\makecell{-1.29\\(0.0***)} &\makecell{2.16\\(0.0***)} &\makecell{-0.15\\(0.0***)} &\makecell{0.33\\(0.0***)} \\
  \hline
  
\end{tabular}
\end{table*}

\end{document}